\documentclass[journal]{IEEEtranTIE}
\usepackage{amsmath,graphicx}
\usepackage{color,hyperref}
\usepackage[table,xcdraw]{xcolor}

\usepackage{subfigure}
\usepackage{times}
\usepackage{epsfig}
\usepackage{graphicx}
\usepackage{amsmath}
\usepackage{amssymb}

\usepackage{algorithm}
\usepackage{algorithmicx}
\usepackage{algpseudocode}
\usepackage{multirow}
\usepackage{booktabs} 
\usepackage{url}
\usepackage{bm}
\usepackage{verbatim}
\usepackage{breakurl}
\usepackage{bbding}
\usepackage{tabu}

\begin{document}
	\title{MT: Multi-Perspective Feature Learning Network for Scene Text Detection}
	
	\author{
		\vskip 1em
		
		Chuang Yang,
		Mulin Chen,
		Yuan Yuan, \emph{Senior Member, IEEE},
		Qi Wang, \emph{Senior Member, IEEE}

			\thanks{The authors are with the School of Computer Science and School
				of Artificial Intelligence, Optics and Electronics (iOPEN), Northwestern
				Polytechnical University, Xi?an 710072, P.R. China. E-mail: cyang113@mail.nwpu.edu.cn, chenmulin@mail.nwpu.edu.cn, y.yuan.ieee@gmail.com, crabwq@gmail.com. }
			\thanks{Qi~Wang is the corresponding author.}
			
	}
	
	\maketitle
	
	\begin{abstract}
		Text detection, the key technology for understanding scene text, has become an attractive research topic. For detecting various scene texts, researchers propose plenty of detectors \textbf{with} different advantages: detection-based models enjoy fast detection speed, and segmentation-based algorithms are not limited by text shapes. However, for most intelligent systems, the detector needs to detect arbitrary-shaped texts with high speed and accuracy simultaneously. Thus, in this study, we design an efficient pipeline named as MT, which can detect adhesive arbitrary-shaped texts with only a single binary mask in the inference stage. This paper presents the contributions on three aspects: (1) a light-weight detection framework is designed to speed up the inference process while keeping high detection accuracy; (2) \textbf{a} multi-perspective feature module is proposed to learn more discriminative representations to \textbf{segment the mask accurately}; (3) a multi-factor constraints IoU minimization loss is introduced for training the proposed \textbf{model}. The effectiveness of MT is evaluated on four real-world scene text datasets, \textbf{and it surpasses all the state-of-the-art competitors to a large extent}.
	\end{abstract}
	
	
	\definecolor{limegreen}{rgb}{0.2, 0.8, 0.2}
	\definecolor{forestgreen}{rgb}{0.13, 0.55, 0.13}
	\definecolor{greenhtml}{rgb}{0.0, 0.5, 0.0}
	
	\section{Introduction}
	%
	%
	%
	%
	Text detection has become increasingly important and popular, which provides fundamental information for some intelligent systems, such as robot navigation, instant translation, and bill recognition systems, etc. Recent progress made in deep convolutional neural networks (CNNs) has greatly boosted the performance of various computer vision tasks like object detection \cite{girshick2014rich,girshick2015fast,ren2015faster,liu2016ssd,redmon2016you,tian2019fcos} and semantic segmentation \cite{long2015fully,badrinarayanan2017segnet,chen2017deeplab,yu2015multi,paszke2016enet,bansal2016pixelnet}, and they further improve the development of text detection. Existing text detectors can be divided into the following two types: the object detection and semantic segmentation-based methods respectively.
	
	\begin{figure}
		\centering
		\subfigure{
			\begin{minipage}[b]{0.98\linewidth}
				\includegraphics[width=1\linewidth]{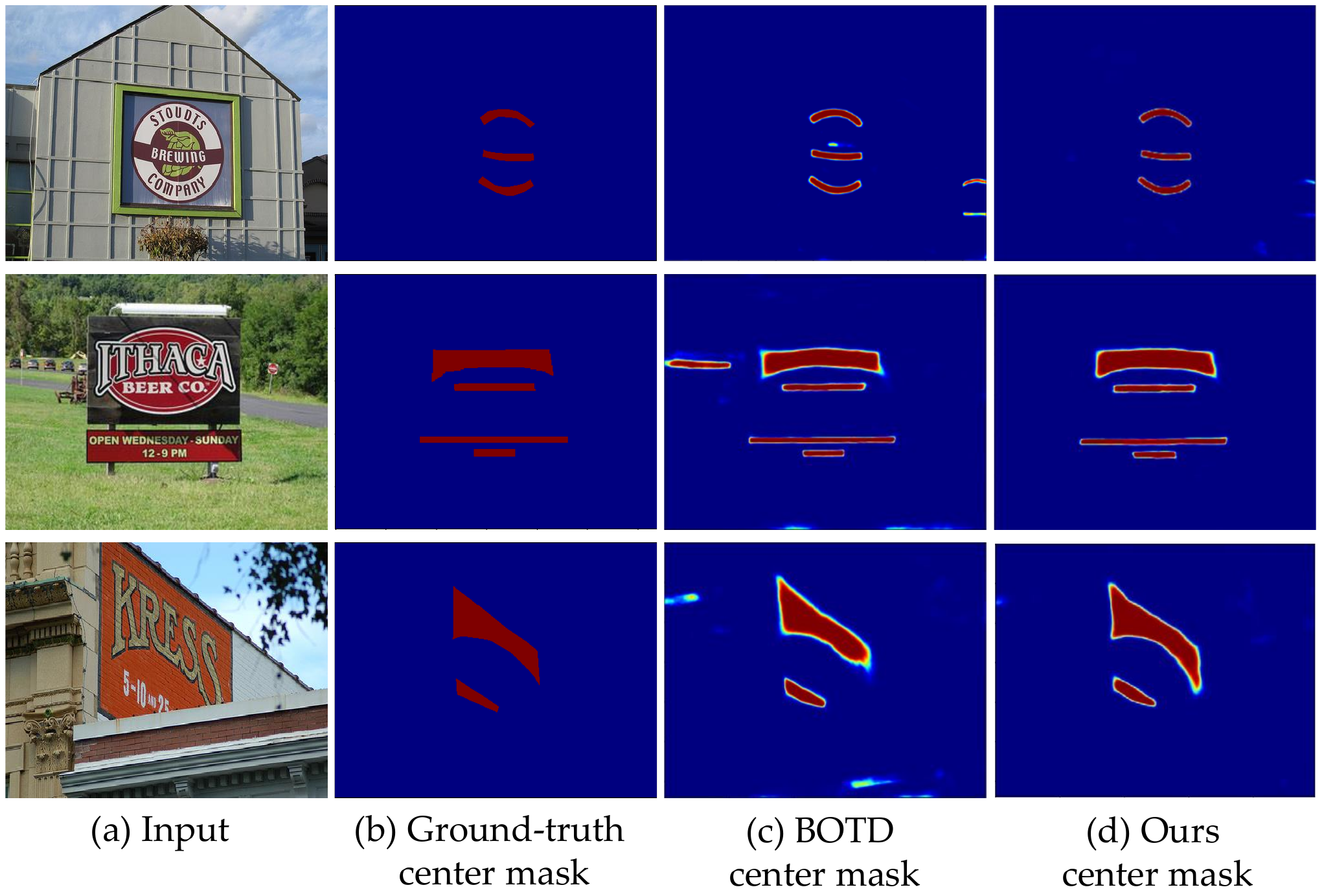}
		\end{minipage}}
		
		\caption{Comparison with MT center mask and BOTD~\cite{yang2020botd} center mask representations. (a) are the input RGB images. (b), (c), and (d) compare the grouth-truth center masks, our proposed MT center masks, and BOTD center masks. The BOTD fails to distinguish text from background (first and second rows) and discriminate center mask from text mask (third row).}
		\label{V1}
	\end{figure}
	
	Object detection-based text detectors enjoy fast detection speed. However, they can not detect the irregular-shaped text effectively. Specifically, these methods predefine a lot of rectangular boxes with different scales and aspect ratios as anchors to match various scene texts at first. Then, the anchor is picked as proposal box when it overlaps a text to a large extent. Next, the offsets between the proposal box and text are predicted. In the end, final detection results are generated by applying the offsets to the corresponding proposal boxes. Since the boxes outputted from the network are quadrilateral, they cannot cover the irregular-shaped text tightly.
	
	To detect arbitrary-shaped texts (such as curved text) accurately, researchers begin to propose segmentation-based text detection methods. Because these algorithms classify image at pixel-level, they can detect arbitrary-shaped texts tightly. However, directly applying these deep CNNs for text detection still suffers from a limitation: adjacent texts can not be separated effectively, which leads to false detection. Considering this problem, researchers introduce various technologies to avoid detecting multiple adhesive texts as one, which also leading to low model efficiency. For example, Wang et al.~\cite{wang2019shape} use text shrink mask to separate those adhesive texts. However, to get the final detection result, they need to generate an extra text mask, and perform a time-consuming pixel-level expansion process. The BOTD~\cite{yang2020botd} models scene texts with Center Mask (CM) and Polar Minimum Distance (PMD). The final detection result can be obtained by applying the PMD to CM directly, which is faster than the pixel-level expansion process. However, since the network includes multiple head branches, the CNNs layers are still not efficient. At the same time, it is difficult to accurately segment CM (as shown in Figure~\ref{V1}~(c)) with the limited supervision information, which deeply influences the detection accuracy.
	
	Considering the aforementioned limitations, how to detect adhesive arbitrary-shaped texts accurately with simple model structure is still under explored. By studying BOTD, we find that the PMD can be computed by CM directly. Thus, we can simplify the CNNs layers by abandoning the PMD head branch. However, it is more difficult to accurately generate CM after removing the PMD head layers. Therefore, we need to provide more supervision information in the training stage to improve the reliability of CM.
	
	In this paper, we design a multi-perspective feature module-based real-time text detector. It can detect arbitrary-shaped scene texts with high detection speed and accuracy, even in the case of text adhesion. In the inference process, the proposed method segments CM with light-weight CNNs layers at first. Then the corresponding PMD is computed according to the CM itself. In the end, the final detection result is obtained by extending CM contour by PMD in the outward direction directly. Moreover, to improve the quality of the generated CM, we introduce a multi-perspective feature module. Specifically, In the training stage, the module encourages MT to learn various text features to accurately discriminate CM from background and text (the representative results as shown in Figure~\ref{V1}~(d)). Importantly, since the module does not appear in the inference stage, which not bring any extra computational cost. The main contributions are as follows. 
	
	\begin{enumerate}
		\item A novel real-time pipeline is designed to detect adhesive arbitrary-shaped texts, which integrates the advantages of both detection and segmentation techniques. In the inference stage, it only includes the following two sub-processes: (1) generating a single CM by light-weight CNNs layers; (2) computing PMD with simple operations and applying it to the corresponding CM directly. In this way, the proposed model saves the computational cost compared with existing methods.
		
		\item A multi-perspective feature module \textbf{is} proposed to learn more discriminative text features. The module includes three sub-modules: (1) Polar minimum distance (PMD) module, (2) Ray distances (RD) module, and (3) GAP module. They encourage the model to recognize text local, edge, and gap features, and further helps the CNNs layers to capture the CM accurately. 
		
		\item A multi-factor constraints IoU maximization loss is introduced for training the multi-perspective feature module. The presented loss function is as insensitive to the text scale, which improves the model robustness for various scales texts. Moreover, the loss function consists of four sub-functions and their error scope are limited to 0$\sim$1, which speeds up the training process and enhances the optimization effectiveness of the offsets compared with existing loss such as the smooth-$l_1$ loss.
	\end{enumerate}
	
	The rest of the paper is organized as follows. Related works about text detection are shown in Section II. The details of our method are presented in Section III. The ablation experiments in Section IV demonstrated the performance of the proposed method. In addition, we compared the MT with its counterparts in Section V. Moreover, the detection results are visualized and detection speed is analyzed in Section VI. Finally, the conclusion of this paper is given in Section VII.

	\section{Related Work}
	\label{Related Work}
	Existing text detection methods mainly contain the following two types: object detection-based methods and semantic segmentation-based methods. In this section, the related works will be reviewed briefly. Additionally, the whole pipeline and text reconstruction process of some representative methods are compared with MT in detail.
	
	\textbf{Object Detection based Methods.} These approaches are inspired by traditional objection detection frameworks, such as Faster-RCNN \cite{ren2015faster}, SSD \cite{liu2016ssd}, and YOLO \cite{redmon2016you}. Liao et al. \cite{liao2016textboxes} changed the size of convolution kernel to extract more effective features. Specifically, they explored the characteristics of text shape and replaced the general 3 $\times$ 3 convolution by 1 $\times$ 5 convolution. Additionally, the authors predefined more anchors with various scales and aspect ratios to match as many scene texts as possible to improve the model recall. Inspired by attention mechanism, He et al. \cite{he2017single} designed a hierarchical inception module. They could get the features with strong representation capacities and reduce the interference information from background. The above methods adopted general detection framework directly with a few adjustments, which only could effectively detect horizontal texts. To get a accurate detection result when detecting multi-oriented texts, they started to redesign model framework. Liao et al. \cite{liao2018textboxes++} designed a angle branch to predicted the rotation angle between the horizontal anchor and multi-oriented text based on their previous work Textboxes~\cite{liao2016textboxes}. The presented framework successfully detect multi-oriented texts with high accuracy. To further improve the performance on the detection of arbitrary-oriented texts, Ma et al. \cite{ma2018arbitrary} proposed some novel anchors to match the declining text. Compared with existing horizontal anchors, the proposed ones have applied a rotation angle to previous ones, which helped the model to cover as many as texts. Additionally, the authors developed rotation region of interesting (RRoI) module to distort the inclined proposals to horizontal boxes, which made the framework to handle rotating regions. Although aforementioned works could detect horizontal and multi-oriented texts with excellent performance, they still suffered from the following limitations: (1) The settings of anchors depended on prior knowledge and the model performance was sensitive to it; (2) Plenty of anchors needed to be predefined to cover various texts, which made the model more complex and slowed down the detection speed largely. Thus, how to effectively detect texts without anchors still under explored. Since anchor-free object detection methods \cite{law2018cornernet,zhu2019feature,kong2020foveabox, li2017multiview} achieved comparable performance in object detection field, researchers started to detect texts with these methods. Zhou et al. \cite{zhou2017east} adopted anchor-free detection framework to predict the offsets between a pixel and the corresponding text border directly, which successfully avoided the influences that the anchor brought to detectors. Although the detection-based text detectors could detect quadrilateral texts, they are failed to accurately find irregular-shaped texts.
	
	\begin{figure*}
		\centering
		\includegraphics[width=1\textwidth]{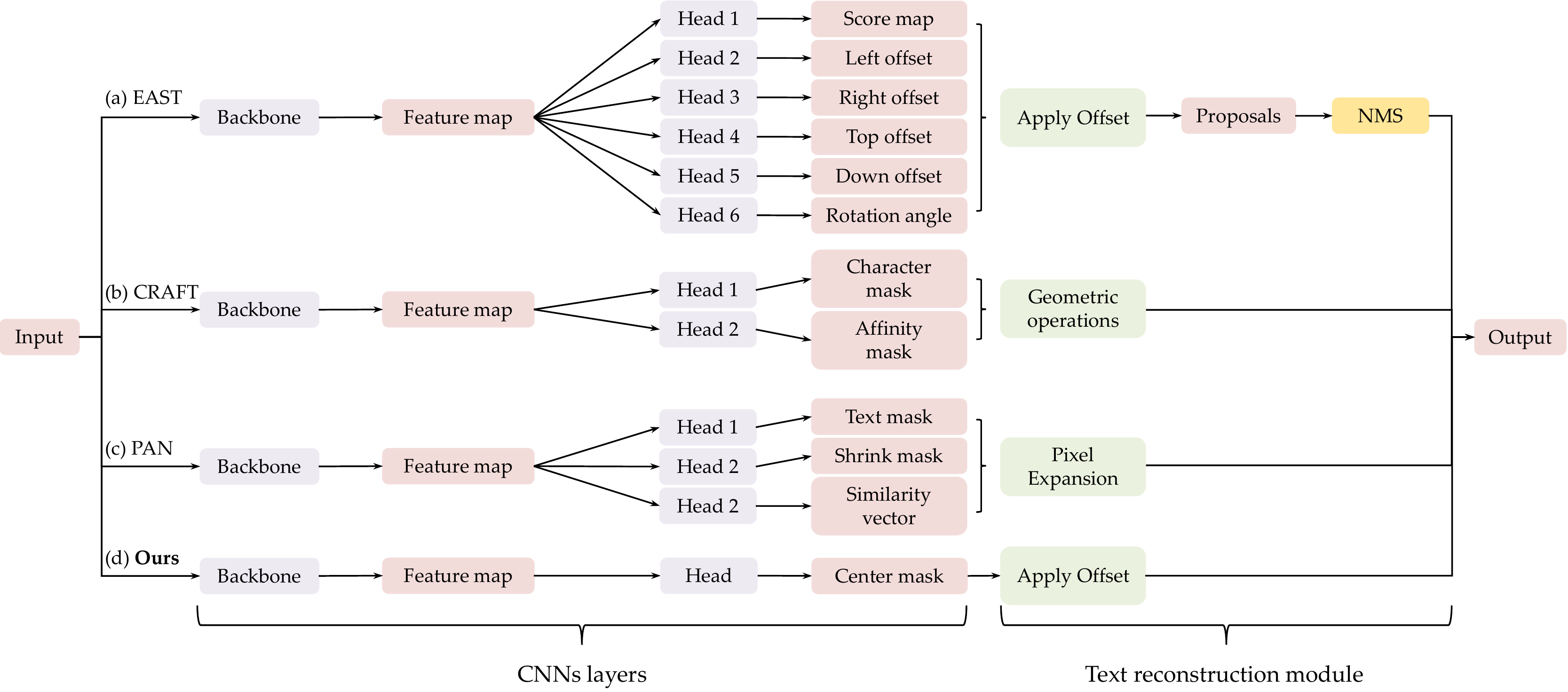}
		\caption{Comparison of pipelines of several representative works on scene text detection: (a) Anchor-free detection framework based multi-oriented shaped text detection pipeline. (b) and (c) are Character-level and word-level segmentation framework based arbitrary-shaped text detection pipelines respectively. (d) Our pipeline, which simplifies the CNNs layers and text reconstruction module greatly.}
		\label{V2}
	\end{figure*}
	
	\textbf{Semantic Segmentation based Methods.} Since Segmentation methods detected objects at pixel-level, they could cover arbitrary-shaped texts tightly. Thus, researchers began to adopt this technology to detect the irregular-shaped text. Long et al. \cite{long2015fully} designed a fully convolutional network (FCN) with multiple CNNs layers. Compared with traditional machine learning-based methods, it enjoyed the following two advantages: (1) FCN outperformed previous general segmentation methods in accuracy to a large extent; (2) It is not limited by the size of input images. The superiorities attracted researchers to propose a series of arbitrary-shaped text detectors based on FCN. Zhang et al. \cite{zhang2016multi} generated multiple salient maps of text and background at first. Then the corresponding text line was computed by salient maps. In the end, final detection result was obtained by connecting text block in the guide of text line. Yao et al. \cite{yao2016scene} segmented word-level and character-level text masks simultaneously, and combined them to further improve the reliability of detect results. Lyu et al. \cite{lyu2018mask} generated the quadrilateral text proposals with detection framework to locate the text roughly. Subsequently, text mask was predicted by the proposals through segmentation methods, which avoided the interferes brought by background. Baek et al. \cite{baek2019character} extended the character-level data through weak supervised network to boost the model performance. Moreover, they proposed 2D Gaussian character label to weak the text edge, which could separate the characters that are close to each other. Following the idea of R-FCN \cite{dai2016r}, Lyu et al. \cite{lyu2018multi} integrated the location information to separate the text into four different regions and combined them to get the final detection results. Although the above methods could accurately detect arbitrary-shaped texts, the text adhesion problem is still effectively avoided. Wang et al. \cite{wang2019shape} shrunk text mask to generate a smaller mask, which could effectively separate adhesive texts and locate them. In addition, the authors segmented text mask and formed the final result by a pixel-level expansion process, which is complicated and time-consuming. Xu et al. \cite{xu2019textfield} used the text direction information to encode the binary text mask and separate adjacent text instances at first. Then, a post-processing based on morphology was applied to get the detection result, which also influenced the detection speed. To speed up the inference process and detect adhesive arbitrary-shaped texts with high accuracy, BOTD~\cite{yang2020botd} modeled the scene texts with Center Mask (CM) and Polar Minimum Distance (PMD). They treated CM and PMD as the anchor and offset respectively. Since the CM is generated by segmentation layers and much tighter than the text boundaries, it can detect arbitrary-shaped texts accurately and the adhesive problem could be avoided naturally. Moreover, they could get the final detection result by extending CM contour by PMD directly, which was faster than pixel-level expansion process.
	
	\textbf{Whole Pipeline Comparison.} A high-level overview of our pipeline is illustrated in Figure~\ref{V2}. For some representative methods (Figure~\ref{V2}~(a-c)), they mostly consist of multiple head branches, components (such as NMS), and complicated text reconstruction steps, which is time consuming and may require extra training data to achieve their best performance. However, as we can see from Figure~\ref{V2}~(d), MT only includes one head layer and a simple text reconstruction steps, which speed up the inference process while keeping high accuracy for the detection of arbitrary-shaped texts. 
	
	Specifically, the whole pipeline can be roughly divided into CNNs layers and text reconstruction module. For the CNNs layers, the proposed method only needs to generate CM with a single head branch, it makes the MT runs faster than other detectors in GPU. For text reconstruction module, our framework inherits the advantages of detection-based detectors that the final detection result is obtained by applying the offset (PMD) on anchor (CM) directly. It is more efficient than the pixel expansion process (Figure~\ref{V2}~(c)) and 
	\begin{figure*}
		\centering
		\includegraphics[width=1\textwidth]{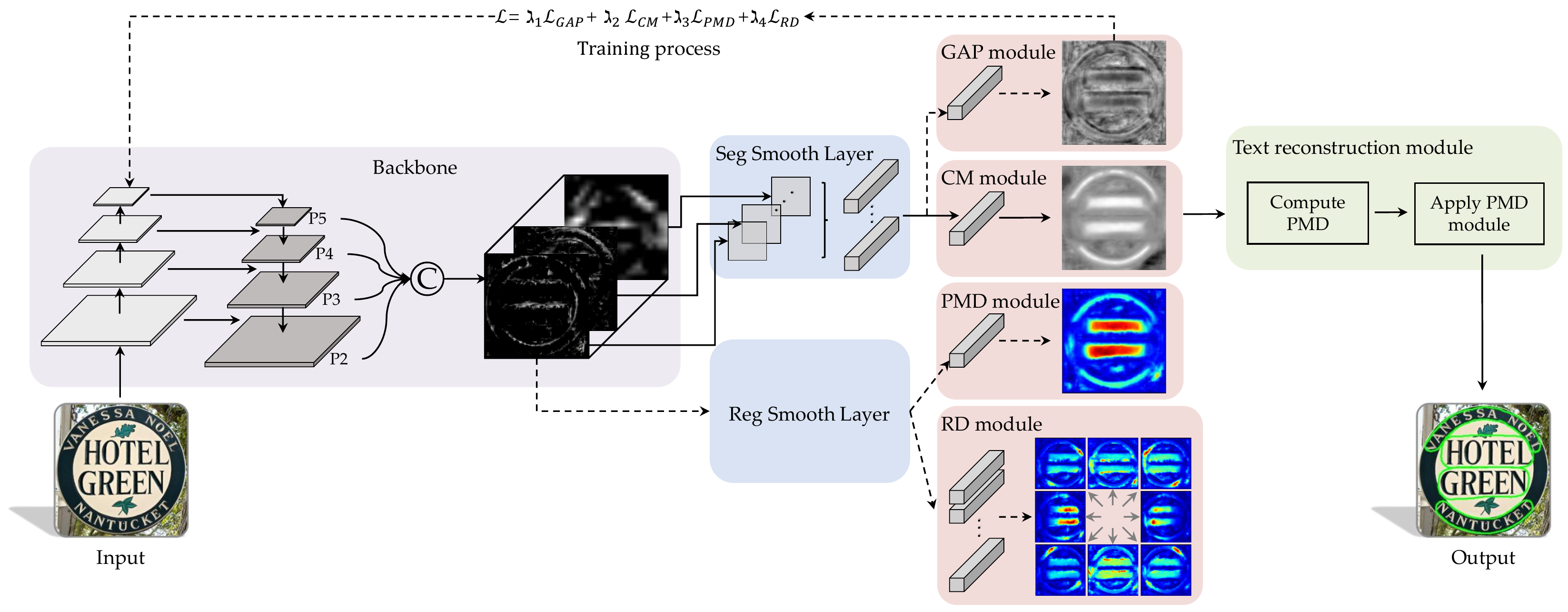}
		\caption{The overall pipeline of MT, which is composed of backbone, double smooth layers, CM and multi-perspective modules, and text reconstruction module. Backbone consist of ResNet and FPN. The two smooth layers enjoy the same structure that includes a deep-wise CNNs layer with 3$\times$3 kernel and a normal CNNs layer with 1$\times$1 kernel. Multi-perspective module have three sub-modules: GAP, PMD, and RD modules. In the text reconstruction module, it needs to compute PMD by CM firstly, and then get the final detection result with apply PMD module.}
		\label{V3}
	\end{figure*}
	the geometric operations (Figure~\ref{V2}~(e)). Moreover, since MT does not need NMS (Figure~\ref{V2}~(a)), the text reconstruction module of the proposed model runs faster than detection-based and segmentation-based algorithms simultaneously in CPU. Therefore, the novel framework not only enjoys the fastest detection speed, but it is easier to achieve all-optimal performance without extra training data, which avoids the heavy dependence of deep learning methods on label data.
	
	\section{Our Method}
	In this section, we introduce the overall structure of the proposed MT at first. Then, the way to simplify the whole pipeline with CM is explained. Next, the multi-perspective feature module is described in detail. In the end, the proposed multi-factor constraints IoU maximization loss and its salient properties are given.
	
	\subsection{Overall Pipeline}
	The whole architecture of the proposed method is shown in Fig. \ref{V3}, which consists of backbone, two smooth layers, a multi-perspective feature module, and text reconstruction module. The backbone is combined by ResNet-18 and FPN to extract the fused features with strong representation and big receptive fields. In addition, for designing a light-weight pipeline, last layers in FPN and the concatenation layer are replaced by group convolution layers. After inputting the RGB image into backbone and getting the concatenate feature map $F_{concat}\in \mathbb{R}^{\left( 512,H/4,W/4 \right)}$ from the concatenation layer ($H$ and $W$ are the height and width of input RGB image). We need further to smooth the $F_{concat}$ to learning the segmentation and regression features ($F_{seg}$ and $F_{reg}$) through two smooth layers that enjoy the same structure respectively:
	\begin{eqnarray}
		F_{seg}=conv_{1\times 1}\left( conv_{3\times 3,group=512}\left( F_{concat} \right) \right), \\
		F_{reg}=conv_{1\times 1}\left( conv_{3\times 3,group=512}\left( F_{concat} \right) \right), 
	\end{eqnarray}
	where $conv_{1\times 1}\left( conv_{3\times 3,group=512}\left( \cdot \right) \right) $ is the CNNs structure of smooth layer, and $\left( conv_{3\times 3,group=512}\left( \cdot \right) \right)$ is a group convolutional layer with 3$\times$3 kernel size and 512 groups. $conv_{1\times 1} \left( \cdot \right)$ is a normal convolutional layer with 1$\times$1 kernel size and 512 channels.
	
	For accurately generating CM by light-weight CNNs layers without extra computational cost, we design a multi-perspective feature module that includes three sub-modules. The three sub-modules and CM module are used to encode the PRCG (Polar minimum distance (PMD), Ray distances (RD), center mask (CM), and GAP) features of text instance and they can be represented as:
	\begin{eqnarray}
		F_{P}=conv_{1\times 1, c=1}\left( F_{reg} \right), \\
		F_{R}=conv_{1\times 1, c=8}\left( F_{reg} \right), \\
		F_{C}=conv_{1\times 1, c=1}\left( F_{seg} \right), \\
		F_{G}=conv_{1\times 1, c=1}\left( F_{seg} \right),
	\end{eqnarray}
	where $F_{P},F_{R},F_{C}$, and $F_{G}$ are the text PMD, RD, CM, and GAP feature maps respectively. $conv_{1\times 1, c=1}(\cdot)$ indicates the convolutional layer with 1$\times$1 kernel size and 1 channel. Therefore, $F_{P},F_{C},F_{G}~\in~\mathbb{R}^{\left( 1,H/4,W/4 \right)} $ and $F_{R}~\in~\mathbb{R}^{\left( 8,H/4,W/4 \right)}$. Since the PMD, RD, and GAP modules (see Figure~\ref{V3} dotted line) are only used for training the network, and the CM module (see Figure~\ref{V3} solid line) used for generating final detection result in the inference stage, the whole inference pipeline can be simplified by abandoning the structure connected with dotted arrows. In this way, the CNNs layers save much time compared with other methods.
	
	After obtaining the CM from CNNs layers, the final detection result can be generated by simple text reconstruction operations. It mainly contains the following two steps: (1) computing PMD by CM; (2) extending CM contour by PMD in outward direction. The text reconstruction process is faster than the pixel-level expansion to a large extent and does not need NMS to filter overlapping results, which further speed up the whole pipeline.
	
	\subsection{Simplify the Pipeline with CM}
	Different from BOTD~\cite{yang2020botd}, our method only need to generate a single CM in the inference stage. Therefore, MT can simplify the whole pipeline by abandoning those CNNs layers that are not related to CM. Moreover, the proposed method inherits the advantages of detection-based methods that it only needs to applying the PMD to CM to get the final detection result, which ensures the superiority in detection speed.
	
	\begin{figure}
		\centering
		\subfigure[The inference process of BOTD]{
			\begin{minipage}[b]{1\linewidth}
				\includegraphics[width=1\linewidth]{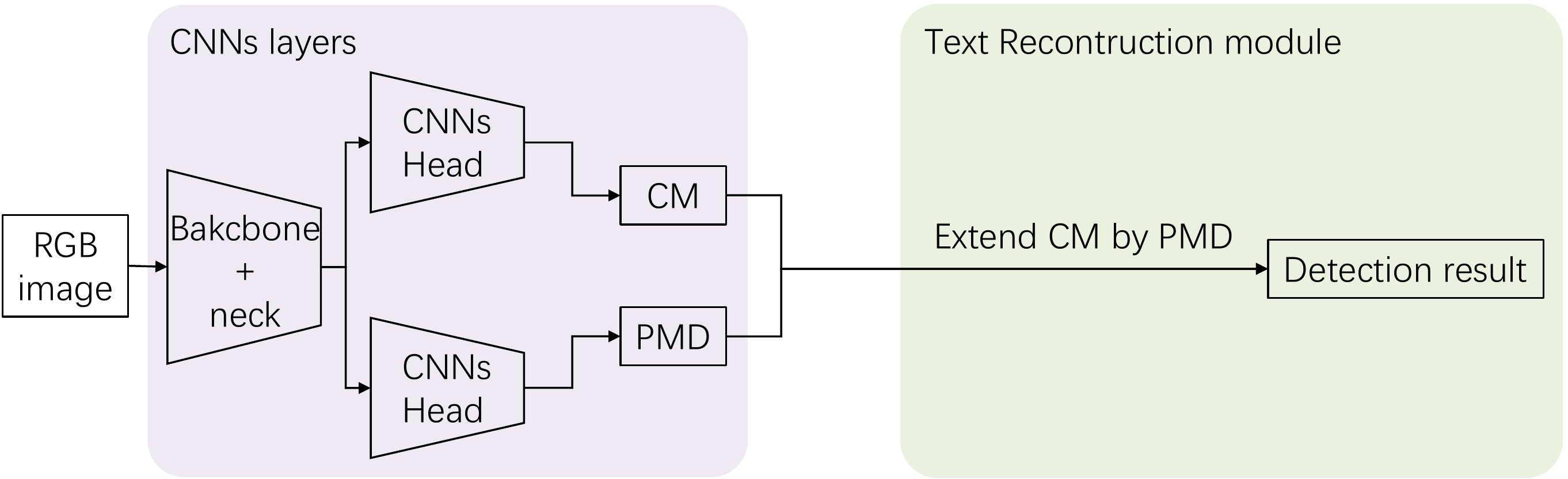}
		\end{minipage}}
		
		\subfigure[The inference process of the proposed MT]{
			\begin{minipage}[b]{1\linewidth}
				\includegraphics[width=1\linewidth]{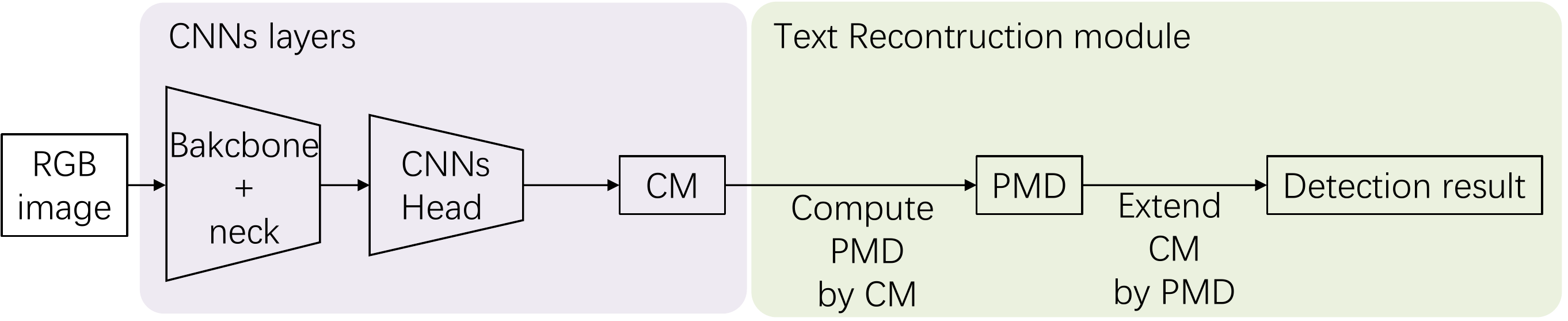}
		\end{minipage}}
		\caption{Comparison of inference processes of BOTD and our method. The whole process is divide into CNNs layers and text reconstruction module.}
		\label{V4}
	\end{figure}
	
\begin{figure}
		\centering
		\includegraphics[width=0.5\textwidth]{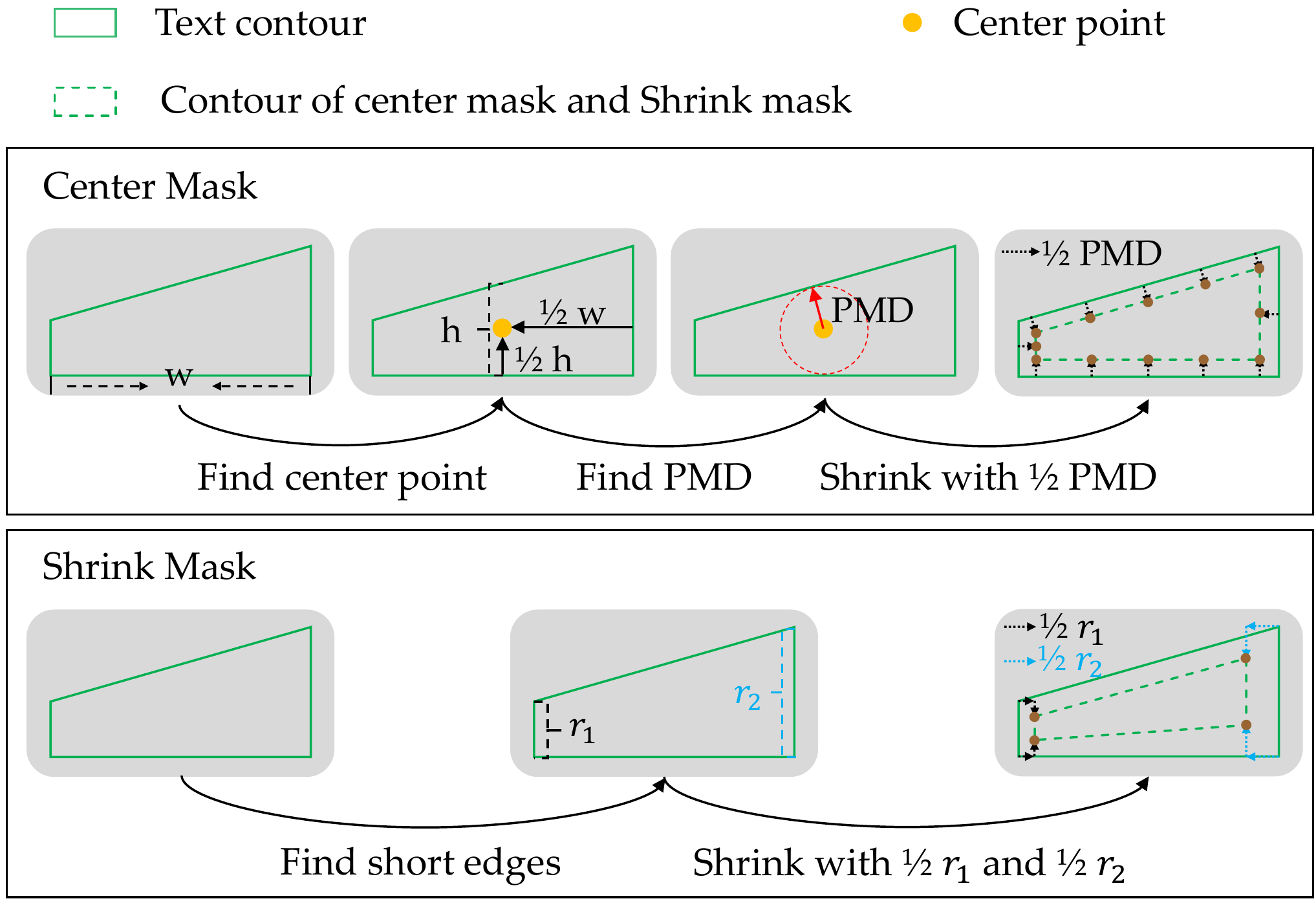}
		\caption{The generation processes of CM and SM, and the essential differences between them.}
		\label{V5}
	\end{figure}

	Here we briefly introduce the CM. It enjoys the same center point with text mask and the distances between CM and text mask contour are equal. Therefore, we only need to segment a single CM and regress the corresponding PMD for arbitrary-shaped text detection, which reduces the complexity of CNNs layers that only includes the two head branches of CM and PMD. Importantly, it is the same as detection-based detectors that the final detection result can be obtained by applying the offset (PMD) on anchor (CM) directly without NMS, which is more efficient than other detection-based and segmentation-based methods. The CM generation process is shown in Figure \ref{V5}~(a), given a trapezoidal text, we need to find the coordinate of center point $p_{cp}$ at first. Supposing the coordinate of text bottom-left corner is $(0,0)$, the $p_{cp}$ can be located by the following formulates:
	\begin{eqnarray}
		p_{cp} = (\frac{1}{2}w,\frac{1}{2}h_{x=\frac{1}{2}w}),
	\end{eqnarray}
	where $w$ is the width of text and $h_{x=\frac{1}{2}w}$ is the height of text when x-coordinate is $\frac{1}{2}w$. After determining the $p_{cp}$, the PMD can be defined as:
	\begin{eqnarray}\label{pmd}
		{\rm PMD}=min(||p_{cp}-P||_2),
	\end{eqnarray}
	where  $\min\left( \cdot \right)$ and $||\cdot ||_2$ indicate the minimum operation and Euclidean distance respectively. $P=\{P_i|i=1,2,...,n\}$ is coordinates set of all points on text contour. Finally, CM is obtained by moving the $P$ inward along the normal direction with $\mu \rm PMD$. $\mu$ is the parameter that determining the moving distance and it is set as 0.5 in this paper. 
	
	As shown in Figure~\ref{V4}~(a), BOTD needs two head branches to generate CM and PMD at first. Then, the final detection result is obtained by applying PMD to CM directly. Inspired by the pipeline of BOTD, the proposed MT also adopts the way to integrating the segmented CM into detection-based framework. In this way, the text reconstruction steps of MT are the same as BOTD and faster than others segmentation-based detectors. Meanwhile, benefit from the advantages of CM, the PMD can be computed with simple operations according to the CM itself. It means that our method only needs to keep the CM head branch in CNNs layers.

	As illustrated in Figure~\ref{V4}~(b), the difference of inference pipelines between the BOTD and our method is that we compute the corresponding PMD by CM, instead of predicting it by generate a heat map. Specifically, in the inference stage, MT needs to get the CM from the simplified CNNs layers at first. Then, obtaining the final detection result with the following two steps: (1) finding the center point of CM and computing the corresponding PMD by CM (the steps as shown in Figure \ref{V5}~(a)); (2) moving all points on CM contour outward along the normal direction with PMD. Although we need more operations to compute PMD compared with BOTD, the computational cost is much less than regressed PMD heat map with CNNs layers.
	
	We also compared CM to the Shrink Mask (SM) that is widely used in previous works \cite{zhou2017east,wang2019shape,wang2019efficient} to locate text and separate adhesive texts. As illustrated in Figure~\ref{V5}~(b), it is generated by moving the text contour inward along the normal direction with different distances. Since the distances between SM and text mask contour are not equal, which means the CNNs layers need to accurately predict all offsets corresponding each point on SM contour at first, and then apply these offsets on SM one by one to get the final detection result. In this way, the model is hard to achieve its best performance and the text reconstruction steps are more time-consuming. Thus, replacing SM by CM is a more efficient way to design a real-time detection framework. 
	
	With the above pipeline, the model efficiency is improved. However, depending on the limited supervision information about the CM, the model is hard to segment CM accurately, which influences the model accuracy. Therefore, we need to provide extra text supervision information to encourage the network to discriminate the CM more effectively.  
	
	\begin{figure}
		\centering
		\includegraphics[width=0.45\textwidth]{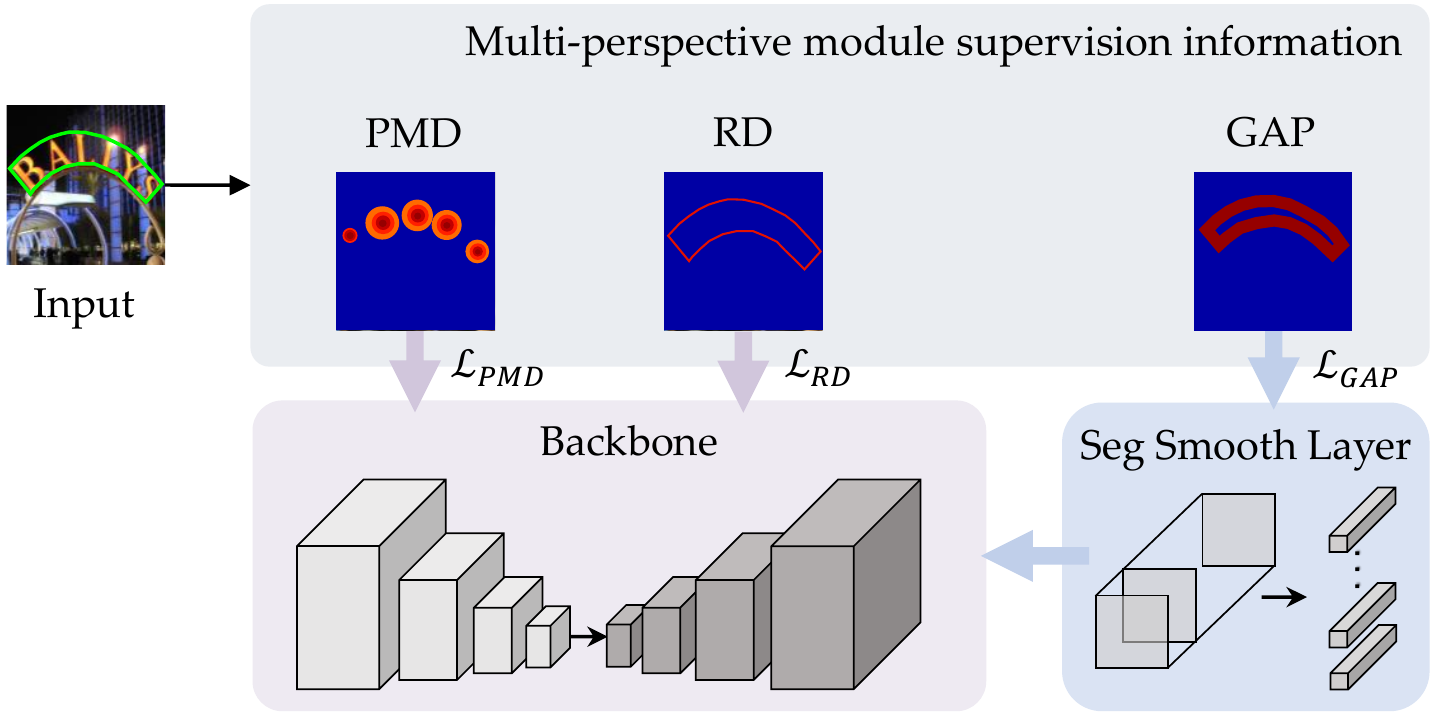}
		\caption{The details of multi-perspective module. PMD and RD modules encourage the backbone to learn the text local feature and text edge feature respectively. The GAP module forces seg smooth layer and backbone to recognize the gap between CM and text contours. ${\cal L}_{\rm RD}, {\cal L}_{\rm PMD},$ and ${\cal L}_{\rm GAP}$ are the RD, PMD, and GAP modules objective functions respectively.}
		\label{V6}
	\end{figure}
	
	\subsection{Multi-Perspective Feature Module}
	Since the light-weight CNNs layers limit the network to learn more discriminative features about text, the CM and PMD are hard to accurately generate. Therefore, we propose the multi-perspective feature module to encourage the MT to recognize the text features from different perspectives and bring no extra computational cost in the inference stage. As shown in Figure~\ref{V6}, the module actually includes PMD, RD, and GAP modules, where PMD and RD modules encourage the model to distinguish text from background and other texts, and GAP module supports the algorithm discriminate CM from the text.
	
\begin{figure*}
		\centering
		\includegraphics[width=1\textwidth]{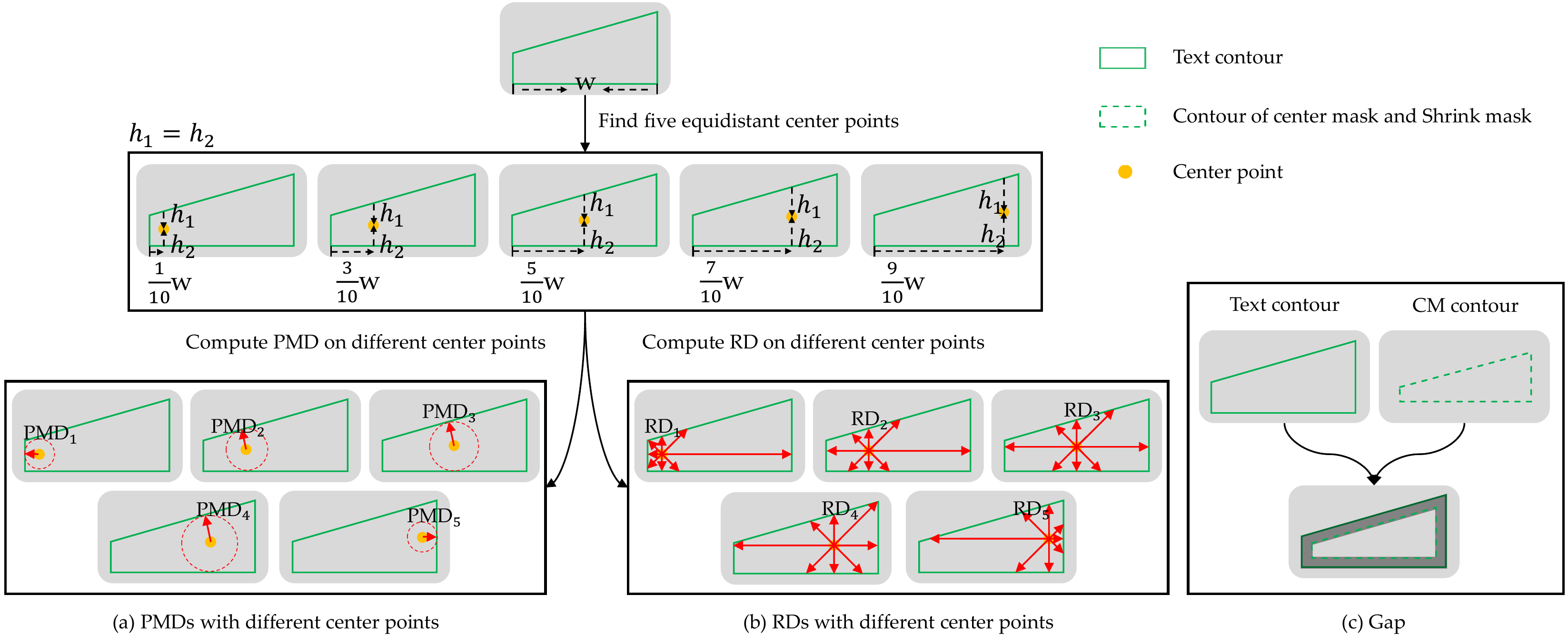}
		\caption{The generation processes of supervision information. (a), (b), and (c) are the supervision information of PMD, RD, and GAP modules respectively. (a) and (b) enjoy the same center points.}
		\label{V7}
	\end{figure*}

	\textbf{PMD module} evaluates the shortest distances between different center points and text contour in 360 directions, which can improve the model ability to recognize the text local features at multiple locations. Therefore, the algorithm can more effectively discriminate the text region from background. As illustrated in Figure~\ref{V3}, the PMD module is a regression branch. Thus, the module only acts on the backbone in the inference process (as we can see from Figure~\ref{V6}). Moreover, Figure~\ref{V7}~(a) vividly shows the generation process of PMD supervision information, given a trapezoidal text, we need to find the coordinates set $P$ of multiple center points:
	
	\begin{eqnarray}
		\begin{gathered}
			P=\{(x,y)\}, \\ 
			y=\frac{1}{2}h_x;~x=\{\frac{1}{10}w,\frac{3}{10}w,\frac{5}{10}w,\frac{7}{10}w,\frac{9}{10}w\},
		\end{gathered}
	\end{eqnarray}
	where $w$ is the width of text, $h_x$ is the text height when x-coordinate is $x$. After locating multiple center points, the label of PMD module can be computed by Eq.~\ref{pmd}. 
	
	\textbf{RD module} \textbf{predicts} multiple offsets between center point and text contour in eight directions (as shown in Figure~\ref{V7}~(b)). Therefore, this module encourages the model to recognize the text edge features when the text region is similar as background or other texts. RD module is the same as PMD module, it is a regression branch and also just acts on the backbone. The label of RD module at a specific center point is generated by:
	
	\begin{eqnarray}
		\begin{gathered}
			RDs=\left\{ ||p_i-p_c||_2 \right\}, \\
			i=1,2,...,8,
		\end{gathered}
	\end{eqnarray}
	where $p_c$ is coordinates of center point that in $P$. $p_i$ are the intersections coordinates between the $p_c$ and text contour in eight directions of up, down, left, right, left top, right top, left down, and right down. 
	Except the aforementioned improvements brought by PMD and RD modules, they also extend the backbone receptive field because the PMD and RD modules are regression tasks. They can enhance the semantic relevance of multiple pixels and further improve the CM accuracy (as shown in Figure~\ref{V1}). Note that, the number of center points is set to 5 in our experiments and it is verified in the ablation study (described in Section~\ref{number}).
	
	\textbf{GAP module} treats the gap between CM and text as background region, which encourages the proposed method to accurately distinguish the CM from text region. As shown in Figure~\ref{V3}, GAP module is the same as CM module that it is a segmentation task. Thus, this module enjoys the same smooth layer with CM module and optimizes the seg smooth layer and backbone simultaneously. The label of GAP module can be generated according to the text mask and Figure~\ref{V7}~(c) vividly shows the corresponding generation process. The supervision information of GAP module is a binary mask between CM and text contours, which can be defined as:
	
	\begin{eqnarray}\label{eqgap}
		{\rm GAP}=1-(M_{text}\cap \left( 1-M_{\rm CM} \right)) ,
	\end{eqnarray}
	where $M_{text}$ and $M_{CM}$ are binary masks of text and CM. Though the three sub-modules encourage the model to learn more discriminative features for segmenting CM, the multi-head branches are hard to optimize and need an effective constraint function to achieve its best performance.
	
	\subsection{Objective Optimization Function}
	In this paper, to optimize the whole pipeline, we propose a multi-factor constraints IoU maximization loss. It contains ${\cal L}_{CM}, {\cal L}_{GAP}, {\cal L}_{PMD},$ and ${\cal L}_{RD}$ four sub objective functions to optimize the corresponding modules in the training stage.
	
	For CM module, the goal is to maximize the IoU between the predicted CM (${\rm CM}_{pred}$) and ground-truth CM (${\rm CM}_{gt}$). In this paper, dice loss is employed to optimize this module, which can enhance the integrity between the optimization objective and the loss function compared with the cross-entropy loss function. The dice loss function is defined as follow:
	\begin{eqnarray}
		{\cal L}_{dice}({{\rm M}_{pred}, {\rm M}_{gt}}) = 1-\frac{2\times| {\rm M}_{pred}\cap {\rm M}_{gt}|+1}{| {\rm M}_{pred} |+| {\rm M}_{gt}|+1},
	\end{eqnarray}
	where ${\rm M}_{pred}$ and ${\rm M}_{gt}$ are the predicted and ground-truth binary masks. The CM loss function based on the ${\cal L}_{dice}$ can be formulated by:
	\begin{eqnarray}
		{\cal L}_{CM} = {\cal L}_{dice}({{\rm CM}_{pred}, {\rm CM}_{gt}}),
	\end{eqnarray}
	
	We find that the gap feature contribute highly to the adhesive text separation and CM edge recognition. Therefore, to supervise the model to effectively extract text gap features by GAP branch, we need to maximize the IoU between the predicted GAP mask (${\rm GAP}_{pred}$) and the corresponding label ${\rm GAP}_{gt}$. Because of the small samples of the text gap feature, only the valid region ${\rm GAP}_{valid}$ in ${\rm GAP}_{pred}$ participate into the training process, and it can be computed by:
	\begin{eqnarray}\label{eqvalid}
		{\rm GAP}_{valid}=\begin{cases}
			{\rm GAP}_{pred},&{\rm GAP}_{gt}=1\\
			0,&		{\rm GAP}_{gt}=0\\
		\end{cases},
	\end{eqnarray}
	
	After obtained the valid region ${\rm GAP}_{valid}$, the GAP loss ${\cal L}_{GAP}$ is defined as follow:
	\begin{eqnarray}
		{\cal L}_{GAP} = {\cal L}_{dice}({{\rm GAP}_{pred}^{valid}, {\rm GAP}_{gt}^{valid}}),
	\end{eqnarray}
	where ${\cal L}_{GAP}$ forces the network to learn the features of valid regions and ignore the `0' regions. The network is not influenced by the background and focuses on the gap regions.
	
	Though segmentation features are effective for representing the scene texts, they may suffer from the ambiguity problem. Thus, in this work, we choose to learn the regression and segmentation features simultaneously to obtain more robust representations for scene text detection. As shown in Fig. \ref{V3}, the regression features are learned by the PMD and RD modules. For improving the model robustness to different scales texts and speed up the training process, ratio loss function is employed to optimize the regression modules in this paper. It treats the difference between the predicted and ground-truth offsets as a ratio, and the value range of it always is 0-1 no matter the text scale is small or big, which makes the model is easy to converge and keep the same sensitivity for all scales of texts. The ratio loss function can be formulated as:
	\begin{eqnarray}
		{\cal L}_{ratio}({{\rm y}_{pred}, {\rm y}_{gt}}) = \log \frac{\max( {\rm y}_{pred}, {\rm y}_{gt})}{\min( {\rm y}_{pred}, {\rm y}_{gt})},
	\end{eqnarray}
	where ${\rm y}_{pred}$ and ${\rm y}_{gt}$ are predicted offset and the corresponding ground-truth respectively. The PMD and RD modules loss functions based on ${\cal L}_{ratio}$ can be defined as:
	\begin{eqnarray}
		{\cal L}_{\rm PMD} = {\cal L}_{ratio}({{\rm PMD}_{pred}, {\rm PMD}_{gt}}),
	\end{eqnarray}
	\begin{eqnarray}
		{\cal L}_{RD} = \frac{1}{m}\sum\nolimits_{k=1}^m{{\cal L}_{ratio}({{\rm RD}_{pred}^{k}, {\rm RD}_{gt}^{k}})}.
	\end{eqnarray}
	where ${\cal L}_{RD}$ considers the offsets in eight directions simultaneously, which enhances the text edge recognition.  
	
	The overall loss function for training the proposed framework is a multi-task loss, which consists of  four terms: 1) the CM loss ${\cal L}_{CM}$; 2) the GAP loss ${\cal L}_{\text{GAP}}$; 3) the PMD loss ${\cal L}_{PMD}$; 4) the RD loss ${\cal L}_{RD}$. It can be represented as
	\begin{equation}
		\begin{aligned}
			{\cal L} = \lambda_1 {\cal L}_{CM} + \lambda_2 {\cal L}_{GAP} + \lambda_3 {\cal L}_{\text{PMD}} + \lambda_4 {\cal L}_{RD}.
		\end{aligned}
	\end{equation}
	where $\lambda_1$, $\lambda_2$, $\lambda_3$ and $\lambda_4$ are balancing weights of loss components for the multi-task loss function.
	
	\section{Experiments}
	Four popular scene text detection datasets are used to evaluate the proposed framework: (1) \textbf{MSRA-TD500} \cite{yao2012detecting} dataset with multi-lingual, arbitrary-oriented and long text lines. It
	\begin{table*}[!t]
		\renewcommand{\arraystretch}{1.3}
		\caption{Detection results on MSRA-TD500 with different settings. Baseline indicate the model only possesses CM module. Center point number means the sampling number of center point in the generation process of supervision information.}
		\setlength{\tabcolsep}{5.75mm}
		\label{ablation}
		\begin{tabular}{l|c|c|c|c|c|c|c|c}
			\hline \hline
			\multirow{2}{*}{\#} & \multirow{2}{*}{Baseline} & \multicolumn{3}{c|}{Multi-perspective module}  & \multirow{2}{*}{\begin{tabular}[c]{@{}c@{}}Center \\ point number\end{tabular}} & \multirow{2}{*}{Precision} & \multirow{2}{*}{Recall} & \multirow{2}{*}{F-measure} \\ \cline{3-5}
			&                           & PMD                       & RD & GAP &                                                                                   &                            &                         &                            \\ \hline
			1                   & \Checkmark                         &                           &    &     &                                                                                   & 87.5                       & 74.6                    & 80.5                       \\ \hline
			2                   & \Checkmark                         & \Checkmark                         &    &     & 1                                                                                 & 85.4                       & 76.3                    & 80.6                       \\ \hline
			3                   & \Checkmark                         & \Checkmark &    &     & 3                                                                                 & 85.5                       & 78.6                    & 81.9                       \\ \hline
			4                   & \Checkmark                         & \Checkmark                         &    &     & 5                                                                                 & 83.8                       & 81.2                    & 82.6 (2.1\%$\uparrow$)                         \\ \hline
			5                   & \Checkmark                         & \Checkmark                         &    &     & 7                                                                                 & 84.5                       & 80.6                    & 82.5                       \\ \hline
			6                   & \Checkmark                         &                           & \Checkmark  &     & 1                                                                                 & 87.8                       & 77.8                    & 82.5                       \\ \hline
			7                   & \Checkmark                         &                           & \Checkmark  &     & 3                                                                                 & 87.5                       & 78.4                    & 82.7                       \\ \hline
			8                   & \Checkmark                         &                           & \Checkmark  &     & 5                                                                                 & 86.7                       & 79.3                    & 82.8 (2.3\%$\uparrow$)                         \\ \hline
			9                   & \Checkmark                         &                           & \Checkmark  &     & 7                                                                                 & 86                         & 79.8                    & 82.8                       \\ \hline
			10                  & \Checkmark                         &                           &    & \Checkmark   & \multicolumn{1}{l|}{}                                                             & 88.4                       & 78                      & 82.9 (2.4\%$\uparrow$)                        \\ \hline
			11                  & \Checkmark                         & \Checkmark                         & \Checkmark  &     & 5                                                                                 & 87                         & 80.8                    & 83.8 (3.3\%$\uparrow$)                         \\ \hline
			12                  & \Checkmark                         & \Checkmark                         & \Checkmark  & \Checkmark   & 5                                                                                 & 90.3                       & 80.5                    & \textbf{85.1} (4.6\%$\uparrow$)                      \\ \hline
		\end{tabular}
	\end{table*}
	contains 700 training images and 200 test images, where the training set includes 400 images from \textbf{HUST-TR400}~\cite{yao2014unified}; (2) \textbf{ICDAR 2015}~\cite{karatzas2015icdar}) dataset is proposed in ICDAR 2015 Robust Reading Competition, and it contains 1500 images, of which 1000 are used to train the model and the remaining  500 are taken as testing set. This dataset contains both horizontal and multi-oriented texts; (3) \textbf{CTW1500}~\cite{yuliang2017detecting} has 1000 training images and 500 testing images. It is a recent challenging dataset for curve text detection and the texts scales vary greatly; (4) \textbf{Total-Text}~\cite{ch2017total} is also a newly-released dataset for curve text detection. This dataset includes horizontal, multi-oriented, and curved text instances simultaneously, and consists of 1255 training images and 300 testing images.
	
	\subsection{Implementation Details}
	We use ResNet18~\cite{he2016deep} and Feature Pyramid Network (FPN)~\cite{lin2017feature} as the model backbone. Data augmentation is used in our work including the following three strategies: (1) random horizontal flipping, (2) random scaling and cropping, (3) random rotating. Note that, MT has not been pre-trained on any external datasets in all following experiments.
	
	In the training stage, Adam \cite{kingma2014adam} is employed with an initial learning rate of 1e-6, and the learning rate is reduced by the ``poly'' learning rate strategy that proposed in~\cite{yu2018bisenet}, in which the initial rate is multiplied by $\left( 1-\frac{iter}{\max \_iter} \right) ^{0.9}$ in all experiments. We train MT with batch size 16 on 2 GPUs for 500 epochs. As for the multi-task training, the parameters ${\lambda_1}$, and ${\lambda_2}$ are set to 0.25 for all experiments.
	
	\subsection{Ablation Study and Discussions}
	To evaluate the proposed framework comprehensively, we do the following ablation studies with ResNet18 and FPN backbone on MSRA-TD500 dataset to explore the effectiveness of different sub-modules and the number of sampling points.
	
	\textbf{The Effectiveness of PMD.}
	The first row in Table~\ref{ablation} displays the model accuracy with baseline, which only includes the CM branch and achieves comparable performance. PMD branch can be used to learning the shortest distance between center point and text contour in 360 directions, which means the model needs to ``look around'' at the position of center point before predicting the PMD. It extends the model receptive field and encourages the network to effectively recognize the text local features.
	We study the effect PMD branch based on the baseline and it (see Table~\ref{ablation}~\#4) can bring over 2\% improvement, which shows the effectiveness of the PMD branch.

	\textbf{The Effectiveness of RD.} RD branch is used to predict the offsets between center point and text contour in eight stationary directions, which improves the model ability to recognize the text edge and discriminate the text from background or other texts. To investigate the effectiveness of RD, we add the RD branch into the baseline. As we can see from Table~\ref{ablation}~\#8, the model with RD branch can make about 2.3\% improvement in F-measure and bring no extra computational cost. This demonstrates that RD branch is useful for detecting scene texts.
	
	\textbf{The Influence of the Number of Center Points.}
	\label{number}
	As shown in Figure~\ref{V7}~(a), MT is sensitive to different local features at multiple center points, which improves the proposed method to recognize the text region accurately. For RD (Figure~\ref{V7}~(b)),
	the model can learn more text edge features when the sampling points number is bigger. Therefore, to better analyze the capability of the proposed PMD and RD branches, the number of center points is varied from 1 to 7 in our experiment. Table~\ref{ablation}~\#2-\#5 are the results of baseline + ``PMD'', we can find that the F-measure on test set keeps rising with the growth of the number and begins to level off when the number ? 7. It is mainly because the light-weight backbone can be effectively trained without a lager amount of samples. The same conclusion (see Table~\ref{ablation}~\#6\~{}\#9) is also applicable to the baseline + ``RD''. Although the F-measures are equal when the number is set to 5 and 7, the latter needs a larger GPU memory in the training process. Therefore, we set the number of center points to 5 in all experiments.
	
	\textbf{The Effectiveness of GAP.}
	We have also evaluated the proposed method performance with ``GAP'' branch based on the baseline. As shown in Table~\ref{ablation}~\#10, the combination model can achieve 82.9\% in F-measure, which outperforms the baseline method 2.4\%. The performance improvement demonstrates that the gap feature is useful for segmenting CM.
	\begin{table*}[]
		\renewcommand\arraystretch{1.3}
		\caption{Comparison with related methods on MSRA-TD500, CTW1500, Total-Text, and ICDAR2015, where `Efficiency' indicates the detection speed and `Ext.? denotes extra training data.}
		\label{comparison}
		\centering
		\setlength{\tabcolsep}{3.5mm}
		\begin{tabular}{c|c|c|c||c|c|c|c||c|c|c|c}
			\hline
			\multirow{2}{*}{Efficiency} & \multirow{2}{*}{Methods} & \multirow{2}{*}{Paper} & \multirow{2}{*}{Ext.}     & \multicolumn{4}{c||}{MSRA-TD500} & \multicolumn{4}{c}{CTW1500}   \\ \cline{5-12} 
			&                          &                        &                           & P(\%)  & R(\%)  & F(\%)  & FPS  & P(\%)  & R(\%)  & F(\%) & FPS  \\ \hline
			\multirow{12}{*}{Low}       & RRD~\cite{liao2018rotation}                      & CVPR'18                & \checkmark & 87.0   & 73.0   & 79.0   & \textbf{10}   & -      & -      & -     & -    \\ \cline{2-12} 
			& TextSnake~\cite{long2018textsnake}                & ECCV'18                & \checkmark & 83.2   & 73.9   & 78.3   & 1.1  & 67.9   & 85.3   & 75.6  & 1.1  \\ \cline{2-12} 
			& SegLink++~\cite{tang2019seglink++}                & PR'19                  & \checkmark & -      & -      & -      & -    & 82.8   & 79.8   & 81.3  & -    \\ \cline{2-12} 
			& CRAFT~\cite{baek2019character}                    & CVPR'19                & \checkmark & \textbf{88.2}   & 78.2   & 82.9   & 8.6  & 86.0   & 81.1   & 83.5  & -    \\ \cline{2-12} 
			& PSE~\cite{wang2019shape}                      & CVPR'19                & $\times$     & -      & -      & -      & -    & 80.6   & 75.6   & 78.0  & 3.9  \\ \cline{2-12} 
			& LOMO~\cite{zhang2019look}                     & CVPR'19                & \checkmark & -      & -      & -      & -    & 85.7   & 76.5   & 80.8  & -    \\ \cline{2-12} 
			& Boundary~\cite{wang2020all}                 & AAAI'20                & \checkmark & -      & -      & -      & -    & -      & -      & -     & -    \\ \cline{2-12} 
			& ContourNet~\cite{wang2020contournet}               & CVPR'20                & $\times$     & -      & -      & -      & -    & 84.1   & \textbf{83.7}   & 83.9  & \textbf{4.5}  \\ \cline{2-12} 
			& DRRG~\cite{zhang2020deep}                     & CVPR'20                & \checkmark & 88.1   & \textbf{82.3}   & \textbf{85.1}   & -    & 85.9   & 83.0   & 84.5  & -    \\ \cline{2-12} 
			& ABCNet~\cite{liu2020abcnet}                   & CVPR'20                & \checkmark & -      & -      & -      & -    & 84.4   & 78.5   & 81.4  & -    \\ \cline{2-12} 
			& TextRay~\cite{wang2020textray}                  & MM'20                  & \checkmark & -      & -      & -      & -    & 82.8   & 80.4   & 81.6  & -    \\ \cline{2-12} 
			& FCENet~\cite{zhu2021fourier}                   & CVPR'21                & $\times$     & -      & -      & -      & -    & \textbf{87.6}   & 83.4   & \textbf{85.5}  & -    \\ \hline
			& EAST~\cite{zhou2017east}                     & CVPR'17                & $\times$     & 87.3   & 67.4   & 76.1   & 13.2 & 78.7   & 49.1   & 60.4  & 21.2 \\ \cline{2-12} 
			\multirow{2}{*}{High}& PAN~\cite{wang2019efficient}                      & ICCV'19                & $\times$     & 80.7   & 77.3   & 78.9   & 30.2 & 84.6   & 77.7   & 81.0  & 39.8 \\ \cline{2-12} 
			& DB~\cite{liao2020real}                      & AAAI'20                & \checkmark     & 90.4   & 76.3   & 82.8   & \textbf{62} & 84.8   & 77.5   & 81.0  & \textbf{55} \\ \cline{2-12} 
			& BOTD~\cite{yang2020botd}                      & Arxiv'20                & $\times$     & 88.3   & 79.2   & 83.5   & 34.8 & 85.7   & 79.6   & 82.5  & 42.6 \\ \cline{2-12} 
			& \textbf{MT}                       & \textbf{Ours}                   & $\times$     & \textbf{90.3}   & \textbf{80.5}   & \textbf{85.1}   & 41.7 & \textbf{86.0}   & \textbf{82.2}   & \textbf{84.1}  & 50.3 \\ \hline \hline
			\multirow{2}{*}{Efficiency} & \multirow{2}{*}{Methods} & \multirow{2}{*}{Paper} & \multirow{2}{*}{Ext.}     & \multicolumn{4}{c||}{Total-Text} & \multicolumn{4}{c}{ICDAR2015} \\ \cline{5-12} 
			&                          &                        &                           & P(\%)  & R(\%)  & F(\%)  & FPS  & P(\%)  & R(\%)  & F(\%) & FPS  \\ \hline
			\multirow{12}{*}{Low}       & RRD~\cite{liao2018rotation}                      & CVPR'18                & \checkmark & -      & -      & -      & -    & 85.6   & 79.0   & 82.2  & 6.5  \\ \cline{2-12} 
			& TextSnake~\cite{long2018textsnake}                & ECCV'18                & \checkmark & 82.7   & 74.5   & 78.4   & -    & 84.9   & 80.4   & 82.6  & 1.1  \\ \cline{2-12} 
			& SegLink++~\cite{tang2019seglink++}                & PR'19                  & \checkmark & -      & -      & -      & -    & 83.7   & 80.3   & 82.0  & \textbf{7.1}  \\ \cline{2-12} 
			& CRAFT~\cite{baek2019character}                    & CVPR'19                & \checkmark & 87.6   & 79.9   & 83.6   & -    & 89.8   & 84.3   & 86.9  & -    \\ \cline{2-12} 
			& PSE~\cite{wang2019shape}                      & CVPR'19                & $\times$     & 81.8   & 75.1   & 78.3   & \textbf{3.9}  & 81.5   & 79.7   & 80.6  & 1.6  \\ \cline{2-12} 
			& LOMO~\cite{zhang2019look}                     & CVPR'19                & \checkmark & 87.6   & 79.3   & 83.3   & -    & 87.8   & 87.6   & \textbf{87.7}  & -    \\ \cline{2-12} 
			& Boundary~\cite{wang2020all}                 & AAAI'20                & \checkmark & 85.2   & 83.5   & 84.3   & -    & 82.2   & \textbf{88.1}   & 85.0  & -    \\ \cline{2-12} 
			& ContourNet~\cite{wang2020contournet}               & CVPR'20                & $\times$     & 86.9   & 83.9   & 85.4   & 3.8  & 87.6   & 86.1   & 86.9  & 3.5  \\ \cline{2-12} 
			& DRRG~\cite{zhang2020deep}                     & CVPR'20                & \checkmark & 86.5   & \textbf{84.9}   & 85.7   & -    & 88.5   & 84.7   & 86.6  & -    \\ \cline{2-12} 
			& ABCNet~\cite{liu2020abcnet}                   & CVPR'20                & \checkmark & 87.9   & 81.3   & 84.5   & -    & -      & -      & -     & -    \\ \cline{2-12} 
			& TextRay~\cite{wang2020textray}                  & MM'20                  & \checkmark & 83.5   & 77.9   & 80.6   & -    & -      & -      & -     & -    \\ \cline{2-12} 
			& FCENet~\cite{zhu2021fourier}                   & CVPR'21                & $\times$     & \textbf{89.3}   & 82.5   & \textbf{85.8}   & -    & \textbf{90.1}   & 82.6   & 86.2  & -    \\ \hline
			& EAST~\cite{zhou2017east}                     & CVPR'17                & $\times$     & 50.0   & 36.2   & 42.0   & -    & 83.6   & 73.5   & 78.2  & 13.2 \\ \cline{2-12} 
			\multirow{2}{*}{High}      & PAN~\cite{wang2019efficient}                      & ICCV'19                & $\times$     & 88.0   & 79.4   & 83.5   & 39.6 & 82.9   & 77.8   & 80.3  & 26.1 \\ \cline{2-12} 
			& DB~\cite{liao2020real}                      & AAAI'20                & \checkmark     & 88.3   & 77.9  & 82.8  & \textbf{50} & 86.8   & 78.4   & 82.3  & \textbf{48} \\ \cline{2-12} 
			& BOTD~\cite{yang2020botd}                      & Arxiv'20                & $\times$     & -   & -   & -   & - & 88.4   & 79.6   & 83.8  & 31.4 \\ \cline{2-12} 
			& \textbf{MT}                       & \textbf{Ours}                   & $\times$     & \textbf{88.5}   & \textbf{81.2}   & \textbf{84.7}   & 49.8 & \textbf{87.3}   & \textbf{81.3}   & \textbf{84.2}  & 34.5 \\ \hline
		\end{tabular}
	\end{table*}
	Since the gap region is also part of text, the model is hard to discriminate CM from text. Therefore, the combination method that exploits both CM and gap features simultaneously to achieve a better detection result than baseline.
	
	\textbf{The Effectiveness of Multi-Tasks.}
	Because the model with aforementioned PMD, RD, and GAP branches can make about 2.1\%, 2.3\%, and 2.4\% improvements respectively in F-measure without any extra computational cost, we verify effectiveness of the composed branches. As we can see from Table~\ref{ablation}~\#11, the composed model with PMD and RD branches simultaneously brings over 1\% improvement compared with the model that includes PMD (see Table~\ref{ablation}~\#4) or RD (see Table~\ref{ablation}~\#8) branch only. Moreover, the performance of the model that is composed of three branches outperforms the methods that consist of baseline and arbitrary single branch in F-measure by at least 2\%. It shows the effectiveness of the combination of PMD, RD, and GAP modules.
	
	\section{Comparison with State-of-the-Art Methods}
	\label{cwem}
	To show more details and give more comprehensive evaluations of the proposed work, we compare our work to other related existing state-of-the-art methods on MSRA-TD500, CTW1500, TotalText, and ICDAR 2015 datasets. In this section, previous SOTA methods are categorized into high-efficiency and low-efficiency according to their detection speed. Specifically, the method is considered as low-efficiency model when the detection speed less than 10 FPS. Otherwise, it is high-efficiency.
	
	\textbf{Evaluation on Multi-language Long Straight Text Benchmark.}
	To test the robustness of MT to multiple languages, we evaluate the proposed method on MSRA-TD500 benchmark. To ensure a fair comparison, the short side of image is set to 736 and evaluating the results with evaluation metrics in~\cite{yuliang2017detecting}.
	
	As shown in Table~\ref{comparison}~MSRA-TD500, MT is superior to all high-efficiency competitors on both detection accuracy and speed. Specifically, the proposed method achieves the F-measure of 85.1\% at an astonishing speed (41.7 FPS). For high-efficiency methods, it outperforms EAST~\cite{zhou2017east}, PAN~\cite{wang2019efficient}, and BOTD~\cite{yang2020botd} by 9\%, 6.2\%, 4\%, and 1.6\% F-measure respectively, and the FPS of our method is 3 times of EAST. Although our method is not as speed as DB~\cite{liao2020real}, MT surpasses it 2.3\% in F-measure. Importantly, the DB improves its detection accuracy by training the model with extra data (SynthText~\cite{gupta2016synthetic}) and replacing the general convolutional layers with the Deformable Convolutional layer~\cite{dai2017deformable}.
	
	For low-efficiency methods, which usually reduce detection speed for improving model accuracy. Such as DGGR~\cite{zhang2020deep}, although MT dose not surpass it in F-measure, our method has a least 10x times faster speed than it. Because the complicated pipeline, DGGR has lower detection speed. Specifically, it not only consists of multiple components, such as two multi-layers head branches, RROI-Align module, and multiple GCN layers, but also needs complex text reconstruction steps to get the final detection result. Therefore, DGGR is hard to converge in the training stage and requires extra training data to obtain its best performance. Different from it, the whole pipeline of MT is very efficient that it only includes one single-layer head branch and simple text reconstruction steps. Thus, the model is easy to converge in the training stage and run faster in the inference process. Importantly, it can achieve the best performance without extra training data, which avoids the heavy dependence of deep learning methods on label data.
	
	The performance on MSRA-TD500 demonstrates the solid superiority and robustness of the proposed MT to detect multi-language long straight text instances. 
	
	\textbf{Evaluation on Curved Text Benchmark.}
	To evaluate the performance of our method for detecting curved and adhesive text instances, the proposed MT is compared with other state-of-the-art methods on CTW1500 and Total-Text datasets respectively. To ensure the fairness of the comparison, we resize the short edge of test images to 640. As we can see from Table~\ref{comparison}~CTW1500 and Total-Text, similar conclusions of the model performance on MSRA-TD500 can be obtained on this two curved text benchmarks. 
	
	On CTW1500 and Total-Text, MT achieves 84.1\% and 84.7\% in F-measure, and 50.3 and 49.8 FPS in detection speed, which surpass all high-efficiency methods to a large extent. For low-efficiency algorithms, although MT is not more accurate than some methods (such as ContourNet, DRRG, and FCENet), our method has a least 12 times faster speed than it. FCENet~\cite{zhu2021fourier} is the same as DGGR, two multi-layers head branches and complex Fourier transformation deeply influence the model detection speed. As for ContourNet~\cite{wang2020contournet}, it extra adds a segmentation network based on the two-stage detection framework, which still includes many extra time-consuming components compared with MT. Although ContourNet achieves 3.5 FPS in detection speed, it is far from MT (49.8 FPS).
	
	The superiority performance on CTW1500 and Total-Text demonstrates the MT can successfully detect adhesive arbitrary-shaped text instances.
	
	\textbf{Evaluation on Multi-Oriented Text Benchmark.} 
	We also verify the MT ability to detect multi-oriented, small, and low-resolution text instances on ICDAR2015 benchmark. During testing, we rescale the short side of input images to 736 like other state-of-the-art methods.
	
	The comparison results are shown in Table~\ref{comparison}, the proposed method outperforms EAST~\cite{zhou2017east}, PAN~\cite{wang2019efficient}, and DB~\cite{liao2020real} by a large margin, i.e., 5.2\%, 3.3\%, and 1.9\% improvements in F-measure, and 3 times and 10 FPS faster than them. For low-efficiency models (such as CRAFT, LOMO, Boundary, and DGGR), they need extra training data to train the complicated framework to achieve the best performance. Therefore, in some scenarios where the images are difficult to collect and the label data is deficient, these methods are hard to achieve their best performance. Importantly, our method has a least 10 times faster speed (34.5 FPS) than them, which means the real-time performance of these methods can not meet the requirements of intelligent systems.
	
	The experiment results on ICDAR 2015 show that the proposed MT can handle the texts with various scales and multi orientations effectively.
	
	\begin{figure*}
		\centering
		\subfigure[MSRA-TD500]{
			\begin{minipage}[b]{0.98\linewidth}
				\includegraphics[width=1\linewidth]{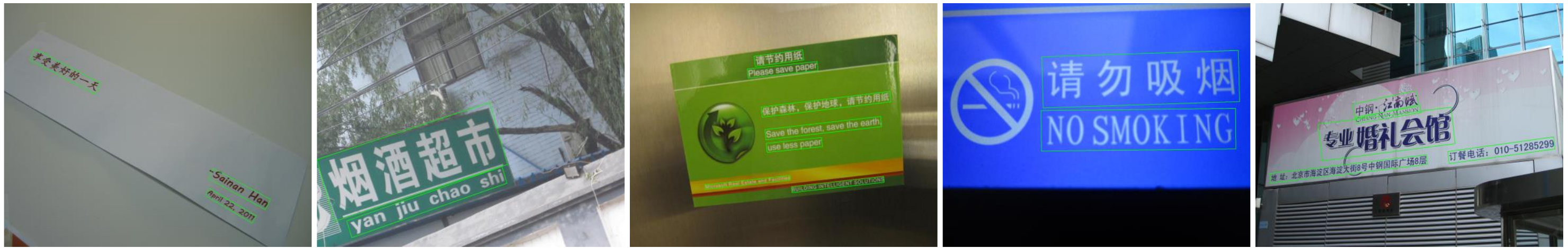}
		\end{minipage}}
		
		\subfigure[CTW1500]{
			\begin{minipage}[b]{0.98\linewidth}
				\includegraphics[width=1\linewidth]{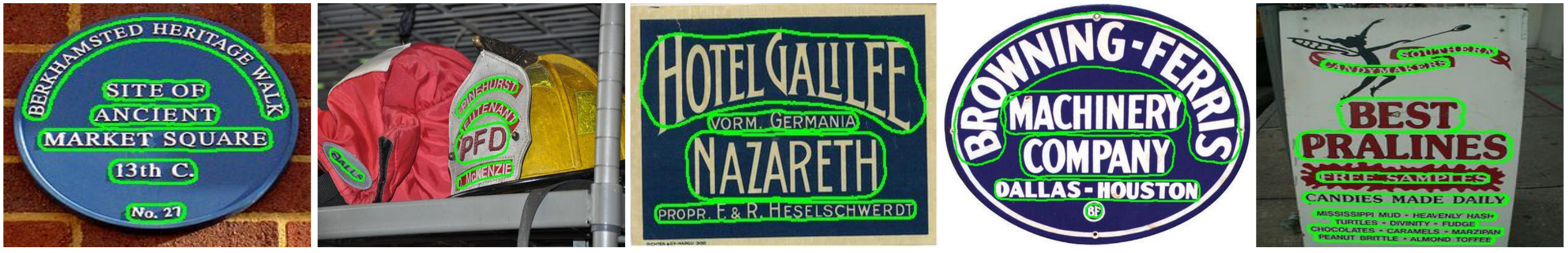}
		\end{minipage}}
		
		\subfigure[Total-Text]{
			\begin{minipage}[b]{0.98\linewidth}
				\includegraphics[width=1\linewidth]{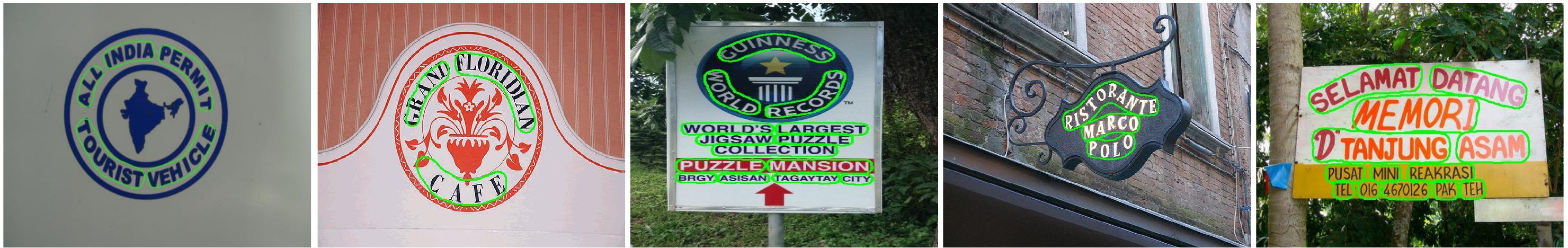}
		\end{minipage}}
		
		\subfigure[ICDAR2015]{
			\begin{minipage}[b]{0.98\linewidth}
				\includegraphics[width=1\linewidth]{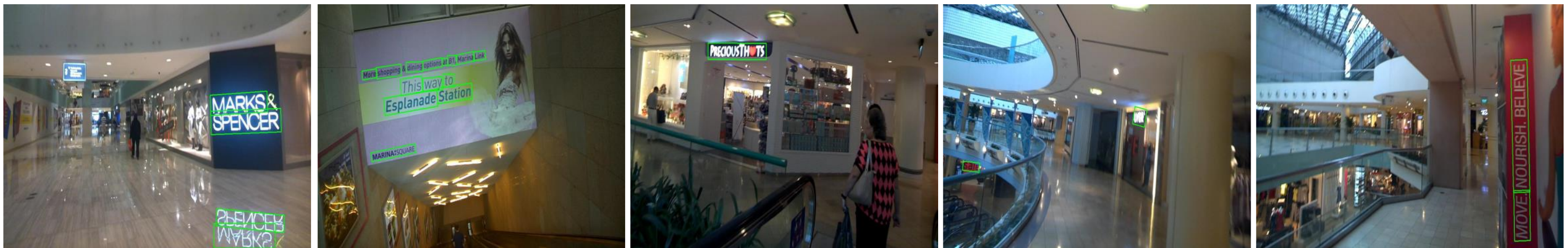}
		\end{minipage}}
		\caption{The visualization of quality detection results on MSRA-TD500, CTW1500, Total-Text, and ICDAR2015 datasets.}
		\label{V8}
	\end{figure*}
	
	\section{Result Visualization and Speed Analysis}
	\textbf{Result Visualization.} The quality detection results on multiple public datasets of MT is shown in Figure~\ref{V8}. From Figure~\ref{V8}~(a), it is verified that the ability of MT to detect multi-language texts simultaneously and the robustness to different scales texts. Figure~\ref{V8}~(b), (c) show the proposed model can effectively avoid the text adhesion problem and detect irregular-shaped scene texts. Moreover, Figure~\ref{V8}~(d) demonstrates the MT is able to accurately detect texts from complicated background.
	
	\begin{table}
		\renewcommand\arraystretch{1.3}
		\label{speed}
		\centering
		\caption{Time consumption of MT on four public benchmarks. The total time
			consists of backbone, segmentation head and post-processing. `Size' and `F' represent the size of short side of input image and F-measure respectively.}
		\setlength{\tabcolsep}{1.62mm}
		\begin{tabular}{|c|c|c|c|c|c|c|}
			\hline
			\multirow{2}{*}{Dataset} & \multirow{2}{*}{Size} & \multirow{2}{*}{F} & \multicolumn{3}{c|}{Time consumption (ms)} & \multirow{2}{*}{FPS} \\ \cline{4-6}
			&                             &                            & Backbone        & Head        & Reconstruct       &                      \\ \hline
			MSRA-TD500               & 736                         & 85.1                       & 11.9            & 9.8         & 2.3        & 41.7                 \\ \hline
			CTW1500                  & 640                         & 84.1                       & 9.7             & 8           & 2.2        & 50.3                 \\ \hline
			Total-Text               & 640                         & 84.7                       & 10              & 7.8         & 2.3        & 49.8                 \\ \hline
			ICDAR2015                & 736                         & 84.2                       & 14.2            & 11.9        & 2.9          & 34.5                 \\ \hline
		\end{tabular}
	\end{table}
	
	\begin{table}
		\renewcommand\arraystretch{1.3}
		\label{FLOP}
		\centering
		\caption{Comparison of computational cost and performance of different real-time detectors. `MAdd' and `FLOPs' indicate Multiply-ACCumulate Operations and Floating Point of Operations respectively.}
		\setlength{\tabcolsep}{1.5mm}
		\begin{tabular}{|c|c|c|c|c|}
			\hline
			\multicolumn{5}{|c|}{CTW1500 (640$\times$640$\times$3)}                           \\ \hline
			Methods & GMAdd         & GFLPOs         & F-measure  & FPS         \\ \hline
			PAN     & 87.21 (33\%$\uparrow$) & 43.67 (+33\%$\uparrow$) & 81.0 (3.1$\downarrow$) & 39.8 (10.5$\downarrow$) \\ \hline
			BOTD    & 62.37 (6\%$\uparrow$)  & 31.25 (+6\%$\uparrow$)  & 82.5 (1.6$\downarrow$) & 42.6 (7.7$\downarrow$)  \\ \hline
			\textbf{Ours}    & \textbf{58.77}         & \textbf{29.44}          & \textbf{84.1}       & \textbf{50.3}        \\ \hline
		\end{tabular}
	\end{table}

	\textbf{Speed Analysis.} We especially analyze the time consumption
	of MT in different stages. As we can see from Table~\ref{speed}, the backbone takes the most time, and the time cost of text reconstruction steps is about a quarter of head. Additionally, the backbone and head will take more time with the increasing of image size. At the same time, it is found that the time cost of text reconstruction steps is not sensitive to the text scale (from Table~\ref{speed}~first and second raws), which means our method can be widely applied to various intelligent systems. However, for traditional segmentation-based methods, the scale and number of text instances influence the time cost of text reconstruction steps deeply simultaneously. Moreover, we compared the computational cost of real-time detectors in Table~\ref{FLOP}. To ensure a fair comparison, under the same setting, we employ ResNet18 as backbone, and resize the size of input image as 640$\times$640$\times$3. As we can see from Table~\ref{FLOP}, our method can reduce 33\% computational cost compared with PAN. At the same time, it brings 3.1\% improvements in F-measure and runs faster than PAN at least 10 FPS on CTW1500. For BOTD, although MT does not reduce the computational cost to a large extent (6\%), The proposed method achieves comparable improvements in detection speed. It mainly because the CM and PMD head branch in BOTD are serial processing mode, GPU needs to compute them one by one. Therefore, after we further simplify the whole pipeline that removing the PMD head leyers, the model can run faster that before a lot. All results in this paper are tested by PyTorch [40] with batch size of 1 on one 1080Ti GPU and one i7-6800k CPU in a single thread.
	
	\section{Conclusion and Future Work}
	In this work, we propose a real-time framework for arbitrary-shaped scene text detection. To overcome the difficulty brought by text shapes, we model the text with CM and PMD, which can cover irregular-shaped texts tightly. The network is simplified by abandoning those CNNs layers that are not related to CM, and introduce a multi-perspective feature module to improve the accuracy without any extra computational cost. Since CM is smaller than the text region, the adhesion problem is avoided naturally instead of various complicated technologies, which saves much computational-cost for text reconstruction module. Moreover, we propose a multi-factor constraints IoU maximization loss to train the whole framework, which not only enjoys a fast convergence process, but also is robust to various scales texts. With the aforementioned advantages, the proposed MT achieves excellent performance in both detection speed and accuracy on multiple public text detection datasets. Experiments demonstrate that the proposed method outperforms all state-of-the-art real-time methods by a large margin when detecting various types of texts.
	
	In the future, we want to design a real-time text spotting model by combining our detector with recognition algorithm, such that an intelligent system can be improved. Additionally, it is desirable to propose a more efficient backbone to extract the feature with strong representation, which can further improve the model accuracy and reduce computational cost.

	\ifCLASSOPTIONcaptionsoff
	\newpage
	\fi

	
	
	\bibliographystyle{IEEEtran}
	\bibliography{egbib}

\begin{thebibliography}{10}
\providecommand{\url}[1]{#1}
\csname url@samestyle\endcsname
\providecommand{\newblock}{\relax}
\providecommand{\bibinfo}[2]{#2}
\providecommand{\BIBentrySTDinterwordspacing}{\spaceskip=0pt\relax}
\providecommand{\BIBentryALTinterwordstretchfactor}{4}
\providecommand{\BIBentryALTinterwordspacing}{\spaceskip=\fontdimen2\font plus
\BIBentryALTinterwordstretchfactor\fontdimen3\font minus
  \fontdimen4\font\relax}
\providecommand{\BIBforeignlanguage}[2]{{%
\expandafter\ifx\csname l@#1\endcsname\relax
\typeout{** WARNING: IEEEtran.bst: No hyphenation pattern has been}%
\typeout{** loaded for the language `#1'. Using the pattern for}%
\typeout{** the default language instead.}%
\else
\language=\csname l@#1\endcsname
\fi
#2}}
\providecommand{\BIBdecl}{\relax}
\BIBdecl

\bibitem{girshick2014rich}
R.~Girshick, J.~Donahue, T.~Darrell, and J.~Malik, ``Rich feature hierarchies
  for accurate object detection and semantic segmentation,'' in
  \emph{Proceedings of the IEEE conference on computer vision and pattern
  recognition}, 2014, pp. 580--587.

\bibitem{girshick2015fast}
R.~Girshick, ``Fast r-cnn,'' in \emph{Proceedings of the IEEE international
  conference on computer vision}, 2015, pp. 1440--1448.

\bibitem{ren2015faster}
S.~Ren, K.~He, R.~Girshick, and J.~Sun, ``Faster r-cnn: Towards real-time
  object detection with region proposal networks,'' in \emph{Proceedings of the
  Neural Information Processing Systems}, 2015, pp. 91--99.

\bibitem{liu2016ssd}
W.~Liu, D.~Anguelov, D.~Erhan, C.~Szegedy, S.~Reed, C.~Fu, and A.~Berg, ``Ssd:
  Single shot multibox detector,'' in \emph{Proceedings of the European
  Conference on Computer Vision}.\hskip 1em plus 0.5em minus 0.4em\relax
  Springer, 2016, pp. 21--37.

\bibitem{redmon2016you}
J.~Redmon, S.~Divvala, R.~Girshick, and A.~Farhadi, ``You only look once:
  Unified, real-time object detection,'' in \emph{Proceedings of the IEEE
  Conference on Computer Vision and Pattern Recognition}, 2016, pp. 779--788.

\bibitem{tian2019fcos}
Z.~Tian, C.~Shen, H.~Chen, and T.~He, ``Fcos: Fully convolutional one-stage
  object detection,'' in \emph{Proceedings of the IEEE International Conference
  on Computer Vision}, 2019, pp. 9627--9636.

\bibitem{long2015fully}
J.~Long, E.~Shelhamer, and T.~Darrell, ``Fully convolutional networks for
  semantic segmentation,'' in \emph{Proceedings of the IEEE Conference on
  Computer Vision and Pattern Recognition}, 2015, pp. 3431--3440.

\bibitem{badrinarayanan2017segnet}
V.~Badrinarayanan, A.~Kendall, and R.~Cipolla, ``Segnet: A deep convolutional
  encoder-decoder architecture for image segmentation,'' \emph{IEEE
  transactions on pattern analysis and machine intelligence}, vol.~39, no.~12,
  pp. 2481--2495, 2017.

\bibitem{chen2017deeplab}
L.-C. Chen, G.~Papandreou, I.~Kokkinos, K.~Murphy, and A.~L. Yuille, ``Deeplab:
  Semantic image segmentation with deep convolutional nets, atrous convolution,
  and fully connected crfs,'' \emph{IEEE transactions on pattern analysis and
  machine intelligence}, vol.~40, no.~4, pp. 834--848, 2017.

\bibitem{yu2015multi}
F.~Yu and V.~Koltun, ``Multi-scale context aggregation by dilated
  convolutions,'' \emph{arXiv preprint arXiv:1511.07122}, 2015.

\bibitem{paszke2016enet}
A.~Paszke, A.~Chaurasia, S.~Kim, and E.~Culurciello, ``Enet: A deep neural
  network architecture for real-time semantic segmentation,'' \emph{arXiv
  preprint arXiv:1606.02147}, 2016.

\bibitem{bansal2016pixelnet}
A.~Bansal, X.~Chen, B.~Russell, A.~Gupta, and D.~Ramanan, ``Pixelnet: Towards a
  general pixel-level architecture,'' \emph{arXiv preprint arXiv:1609.06694},
  2016.

\bibitem{yang2020botd}
C.~Yang, Z.~Xiong, M.~Chen, Q.~Wang, and X.~Li, ``Botd: Bold outline text
  detector,'' \emph{arXiv preprint arXiv:2011.14714}, 2020.

\bibitem{wang2019shape}
W.~Wang, E.~Xie, X.~Li, W.~Hou, T.~Lu, G.~Yu, and S.~Shao, ``Shape robust text
  detection with progressive scale expansion network,'' in \emph{Proceedings of
  the IEEE Conference on Computer Vision and Pattern Recognition}, 2019, pp.
  9336--9345.

\bibitem{liao2016textboxes}
M.~Liao, B.~Shi, X.~Bai, X.~Wang, and W.~Liu, ``Textboxes: A fast text detector
  with a single deep neural network,'' \emph{arXiv preprint arXiv:1611.06779},
  2016.

\bibitem{he2017single}
P.~He, W.~Huang, T.~He, Q.~Zhu, Y.~Qiao, and X.~Li, ``Single shot text detector
  with regional attention,'' in \emph{Proceedings of the IEEE International
  Conference on Computer Vision}, 2017, pp. 3047--3055.

\bibitem{liao2018textboxes++}
M.~Liao, B.~Shi, and X.~Bai, ``Textboxes++: A single-shot oriented scene text
  detector,'' \emph{IEEE Transactions on Image Processing}, vol.~27, no.~8, pp.
  3676--3690, 2018.

\bibitem{ma2018arbitrary}
J.~Ma, W.~Shao, H.~Ye, L.~Wang, H.~Wang, Y.~Zheng, and X.~Xue,
  ``Arbitrary-oriented scene text detection via rotation proposals,''
  \emph{IEEE Transactions on Multimedia}, vol.~20, no.~11, pp. 3111--3122,
  2018.

\bibitem{law2018cornernet}
H.~Law and J.~Deng, ``Cornernet: Detecting objects as paired keypoints,'' in
  \emph{Proceedings of the European Conference on Computer Vision}, 2018, pp.
  734--750.

\bibitem{zhu2019feature}
C.~Zhu, Y.~He, and M.~Savvides, ``Feature selective anchor-free module for
  single-shot object detection,'' in \emph{Proceedings of the IEEE Conference
  on Computer Vision and Pattern Recognition}, 2019, pp. 840--849.

\bibitem{kong2020foveabox}
T.~Kong, F.~Sun, H.~Liu, Y.~Jiang, L.~Li, and J.~Shi, ``Foveabox: Beyound
  anchor-based object detection,'' \emph{IEEE Transactions on Image
  Processing}, vol.~29, pp. 7389--7398, 2020.

\bibitem{li2017multiview}
X.~Li, M.~Chen, F.~Nie, and Q.~Wang, ``A multiview-based parameter free
  framework for group detection,'' in \emph{Proceedings of the AAAI Conference
  on Artificial Intelligence}, vol.~31, no.~1, 2017.

\bibitem{zhou2017east}
X.~Zhou, C.~Yao, H.~Wen, Y.~Wang, S.~Zhou, W.~He, and J.~Liang, ``East: an
  efficient and accurate scene text detector,'' in \emph{Proceedings of the
  IEEE Conference on Computer Vision and Pattern Recognition}, 2017, pp.
  5551--5560.

\bibitem{zhang2016multi}
Z.~Zhang, C.~Zhang, W.~Shen, C.~Yao, W.~Liu, and X.~Bai, ``Multi-oriented text
  detection with fully convolutional networks,'' in \emph{Proceedings of the
  IEEE Conference on Computer Vision and Pattern Recognition}, 2016, pp.
  4159--4167.

\bibitem{yao2016scene}
C.~Yao, X.~Bai, N.~Sang, X.~Zhou, S.~Zhou, and Z.~Cao, ``Scene text detection
  via holistic, multi-channel prediction,'' \emph{arXiv preprint
  arXiv:1606.09002}, 2016.

\bibitem{lyu2018mask}
P.~Lyu, M.~Liao, C.~Yao, W.~Wu, and X.~Bai, ``Mask textspotter: An end-to-end
  trainable neural network for spotting text with arbitrary shapes,'' in
  \emph{Proceedings of the European Conference on Computer Vision (ECCV)},
  2018, pp. 67--83.

\bibitem{baek2019character}
Y.~Baek, B.~Lee, D.~Han, S.~Yun, and H.~Lee, ``Character region awareness for
  text detection,'' in \emph{Proceedings of the IEEE Conference on Computer
  Vision and Pattern Recognition}, 2019, pp. 9365--9374.

\bibitem{dai2016r}
J.~Dai, Y.~Li, K.~He, and J.~Sun, ``R-fcn: Object detection via region-based
  fully convolutional networks,'' \emph{arXiv preprint arXiv:1605.06409}, 2016.

\bibitem{lyu2018multi}
P.~Lyu, C.~Yao, W.~Wu, S.~Yan, and X.~Bai, ``Multi-oriented scene text
  detection via corner localization and region segmentation,'' in
  \emph{Proceedings of the IEEE Conference on Computer Vision and Pattern
  Recognition}, 2018, pp. 7553--7563.

\bibitem{xu2019textfield}
Y.~Xu, Y.~Wang, W.~Zhou, Y.~Wang, Z.~Yang, and X.~Bai, ``Textfield: Learning a
  deep direction field for irregular scene text detection,'' \emph{IEEE
  Transactions on Image Processing}, vol.~28, no.~11, pp. 5566--5579, 2019.

\bibitem{wang2019efficient}
W.~Wang, E.~Xie, X.~Song, Y.~Zang, W.~Wang, T.~Lu, G.~Yu, and C.~Shen,
  ``Efficient and accurate arbitrary-shaped text detection with pixel
  aggregation network,'' in \emph{Proceedings of the IEEE International
  Conference on Computer Vision}, 2019, pp. 8440--8449.

\bibitem{yao2012detecting}
C.~Yao, X.~Bai, W.~Liu, Y.~Ma, and Z.~Tu, ``Detecting texts of arbitrary
  orientations in natural images,'' in \emph{2012 IEEE conference on computer
  vision and pattern recognition}.\hskip 1em plus 0.5em minus 0.4em\relax IEEE,
  2012, pp. 1083--1090.

\bibitem{yao2014unified}
C.~Yao, X.~Bai, and W.~Liu, ``A unified framework for multioriented text
  detection and recognition,'' \emph{IEEE Transactions on Image Processing},
  vol.~23, no.~11, pp. 4737--4749, 2014.

\bibitem{karatzas2015icdar}
D.~Karatzas, L.~Gomez, A.~Nicolaou, S.~Ghosh, A.~Bagdanov, M.~Iwamura,
  J.~Matas, L.~Neumann, V.~Chandrasekhar, and S.~Lu, ``Icdar 2015 competition
  on robust reading,'' in \emph{Proceedings of the International Conference on
  Document Analysis and Recognition}.\hskip 1em plus 0.5em minus 0.4em\relax
  IEEE, 2015, pp. 1156--1160.

\bibitem{yuliang2017detecting}
Y.~Liu, L.~Jin, S.~Zhang, and S.~Zhang, ``Detecting curve text in the wild: New
  dataset and new solution,'' \emph{arXiv preprint arXiv:1712.02170}, 2017.

\bibitem{ch2017total}
C.~K. Ch'ng and C.~S. Chan, ``Total-text: A comprehensive dataset for scene
  text detection and recognition,'' in \emph{Proceedings of the International
  Conference on Document Analysis and Recognition}, vol.~1.\hskip 1em plus
  0.5em minus 0.4em\relax IEEE, 2017, pp. 935--942.

\bibitem{he2016deep}
K.~He, X.~Zhang, S.~Ren, and J.~Sun, ``Deep residual learning for image
  recognition,'' in \emph{Proceedings of the IEEE Conference on Computer Vision
  and Pattern Recognition}, 2016, pp. 770--778.

\bibitem{lin2017feature}
T.-Y. Lin, P.~Doll{\'a}r, R.~Girshick, K.~He, B.~Hariharan, and S.~Belongie,
  ``Feature pyramid networks for object detection,'' in \emph{Proceedings of
  the IEEE conference on computer vision and pattern recognition}, 2017, pp.
  2117--2125.

\bibitem{kingma2014adam}
D.~P. Kingma and J.~Ba, ``Adam: A method for stochastic optimization,''
  \emph{arXiv preprint arXiv:1412.6980}, 2014.

\bibitem{yu2018bisenet}
C.~Yu, J.~Wang, C.~Peng, C.~Gao, G.~Yu, and N.~Sang, ``Bisenet: Bilateral
  segmentation network for real-time semantic segmentation,'' in
  \emph{Proceedings of the European conference on computer vision (ECCV)},
  2018, pp. 325--341.

\bibitem{liao2018rotation}
M.~Liao, Z.~Zhu, B.~Shi, G.~Xia, and X.~Bai, ``Rotation-sensitive regression
  for oriented scene text detection,'' in \emph{Proceedings of the IEEE
  Conference on Computer Vision and Pattern Recognition}, 2018, pp. 5909--5918.

\bibitem{long2018textsnake}
S.~Long, J.~Ruan, W.~Zhang, X.~He, W.~Wu, and C.~Yao, ``Textsnake: A flexible
  representation for detecting text of arbitrary shapes,'' in \emph{Proceedings
  of the European Conference on Computer Vision}, 2018, pp. 20--36.

\bibitem{tang2019seglink++}
J.~Tang, Z.~Yang, Y.~Wang, Q.~Zheng, Y.~Xu, and X.~Bai, ``Seglink++: Detecting
  dense and arbitrary-shaped scene text by instance-aware component grouping,''
  \emph{Pattern recognition}, vol.~96, p. 106954, 2019.

\bibitem{zhang2019look}
C.~Zhang, B.~Liang, Z.~Huang, M.~En, J.~Han, E.~Ding, and X.~Ding, ``Look more
  than once: An accurate detector for text of arbitrary shapes,'' in
  \emph{Proceedings of the IEEE/CVF Conference on Computer Vision and Pattern
  Recognition}, 2019, pp. 10\,552--10\,561.

\bibitem{wang2020all}
H.~Wang, P.~Lu, H.~Zhang, M.~Yang, X.~Bai, Y.~Xu, M.~He, Y.~Wang, and W.~Liu,
  ``All you need is boundary: Toward arbitrary-shaped text spotting,'' in
  \emph{Proceedings of the AAAI Conference on Artificial Intelligence},
  vol.~34, no.~07, 2020, pp. 12\,160--12\,167.

\bibitem{wang2020contournet}
Y.~Wang, H.~Xie, Z.~Zha, M.~Xing, Z.~Fu, and Y.~Zhang, ``Contournet: Taking a
  further step toward accurate arbitrary-shaped scene text detection,'' in
  \emph{Proceedings of the IEEE Conference on Computer Vision and Pattern
  Recognition}, 2020, pp. 11\,753--11\,762.

\bibitem{zhang2020deep}
S.-X. Zhang, X.~Zhu, J.-B. Hou, C.~Liu, C.~Yang, H.~Wang, and X.-C. Yin, ``Deep
  relational reasoning graph network for arbitrary shape text detection,'' in
  \emph{Proceedings of the IEEE/CVF Conference on Computer Vision and Pattern
  Recognition}, 2020, pp. 9699--9708.

\bibitem{liu2020abcnet}
Y.~Liu, H.~Chen, C.~Shen, T.~He, L.~Jin, and L.~Wang, ``Abcnet: Real-time scene
  text spotting with adaptive bezier-curve network,'' in \emph{Proceedings of
  the IEEE/CVF Conference on Computer Vision and Pattern Recognition}, 2020,
  pp. 9809--9818.

\bibitem{wang2020textray}
F.~Wang, Y.~Chen, F.~Wu, and X.~Li, ``Textray: Contour-based geometric modeling
  for arbitrary-shaped scene text detection,'' in \emph{Proceedings of the 28th
  ACM International Conference on Multimedia}, 2020, pp. 111--119.

\bibitem{zhu2021fourier}
Y.~Zhu, J.~Chen, L.~Liang, Z.~Kuang, L.~Jin, and W.~Zhang, ``Fourier contour
  embedding for arbitrary-shaped text detection,'' \emph{arXiv preprint
  arXiv:2104.10442}, 2021.

\bibitem{liao2020real}
M.~Liao, Z.~Wan, C.~Yao, K.~Chen, and X.~Bai, ``Real-time scene text detection
  with differentiable binarization.'' in \emph{Proceedings of the AAAI
  Conference on Artificial Intelligence}, 2020, pp. 11\,474--11\,481.

\bibitem{gupta2016synthetic}
A.~Gupta, A.~Vedaldi, and A.~Zisserman, ``Synthetic data for text localisation
  in natural images,'' in \emph{Proceedings of the IEEE conference on computer
  vision and pattern recognition}, 2016, pp. 2315--2324.

\bibitem{dai2017deformable}
J.~Dai, H.~Qi, Y.~Xiong, Y.~Li, G.~Zhang, H.~Hu, and Y.~Wei, ``Deformable
  convolutional networks,'' in \emph{Proceedings of the IEEE international
  conference on computer vision}, 2017, pp. 764--773.

\end{thebibliography}

\end{document}